# Multi-model approach for autonomous driving: A comprehensive study on traffic sign-, vehicle- and lane-detection and behavioral cloning

Kanishkha Jaisankar[1], Pranav M. Pawar[2], Diana Susane Joseph[3], Raja Muthalagu[4], Mithun Mukherjee[5], Dnyaneshawar Mantri[6], Ramjee Prasad[7]
[1,2,3,4,5]Department of Computer Science,
Birla Institute of Technology and Science Pilani,
Dubai Campus, Dubai, UAE.
[6]Sinhgad Institute of Technology, Lonavala
[7]CTIF Global Capsule, Denmark
[1] f20190072d@alumni.bits-pilani.ac.in , [3]diclmi19@gmail.com
[2,4,5]{pranav, raja.m, mithun}@dubai.bits-pilani.ac.in
[6]dsmantri.sit@sinhgad.edu, [7]ramjee@btech.au.dk

***Abstract*** – ***Deep learning and computer vision techniques have become increasingly important in the development of self-driving cars. These techniques play a crucial role in enabling self-driving cars to perceive and understand their surroundings, allowing them to safely navigate and make decisions in real-time. Using Neural Networks self-driving cars can accurately identify and classify objects such as pedestrians, other vehicles, and traffic signals. Using deep learning and analyzing data from sensors such as cameras and radar, self-driving cars can predict the likely movement of other objects and plan their own actions accordingly. In this study, a novel approach to enhance the performance of self-driving cars by using pre-trained and custom-made neural networks for key tasks, including traffic sign classification, vehicle detection, lane detection, and behavioral cloning is provided. The methodology integrates several innovative techniques, such as geometric and color transformations for data augmentation, image normalization, and transfer learning for feature extraction. These techniques are applied to diverse datasets, including the German Traffic Sign Recognition Benchmark (GTSRB), road and lane segmentation datasets, vehicle detection datasets, and data collected using the Udacity self-driving car simulator to evaluate the model efficacy. The primary objective of the work is to review the state-of-the-art in deep learning and computer vision for self-driving cars. The findings of the work are effective in solving various challenges related to self-driving cars like traffic sign classification, lane prediction, vehicle detection, and behavioral cloning, and provide valuable insights into improving the robustness and reliability of autonomous systems, paving the way for future research and deployment of safer and more efficient self-driving technologies.***

## 1. Introduction

Self-driving cars, also known as autonomous vehicles, have the potential to revolutionize transportation by providing a safer, more efficient, and more convenient alternative to traditional human-driven vehicles. According to a report from the National Highway Traffic Safety Administration (NHTSA), 94% of car accidents in the United States are caused by human error, such as distracted, drunk driving, poor decision-making, or impaired driving [1]. By eliminating the need for a human driver, self-driving cars have the potential to significantly reduce the number of accidents and fatalities on the roads and save lives.

Additionally, self-driving cars could allow people who are unable to drive due to age or disability to regain their independence and mobility.

To achieve this vision, self-driving cars rely on a variety of advanced sensors and technologies to navigate and understand their surroundings. These sensors include cameras, LIDAR (Light Detection and Ranging), radar, and GPS (Global Positioning System). The data from these sensors is processed by advanced computer systems, including GPUs (Graphics Processing Units) and TPUs (Tensor Processing Units), to create a detailed understanding of the environment around the car.

One key area that has received significant attention in the development of self-driving cars is the use of advanced deep learning and computer vision techniques to enable the car to perceive and understand its environment. Deep learning is a type of machine learning that involves training artificial neural networks on large datasets to recognize patterns and make decisions. Convolutional neural networks (CNNs) are a type of deep learning model that is particularly well-suited for image recognition tasks. In self-driving cars, CNNs can be used to identify and classify objects such as pedestrians, other vehicles, and traffic signals with high accuracy and speed [2].

Computer vision, the field of study focused on enabling computers to interpret and understand visual data from the world, is also critical for self-driving cars. OpenCV (Open-Source Computer Vision) is a widely used library for implementing computer vision techniques, including object detection and tracking. In self-driving cars, these techniques can be used to identify and track objects in the car's environment, allowing it to make informed decisions about how to navigate safely and efficiently. Additionally, advanced computer vision techniques such as 3D scene reconstruction, optical flow estimation, and semantic segmentation are being explored to provide an even more detailed and accurate understanding of the environment [3].

Numerous studies have employed various techniques, such as object detection and deep learning, to advance the development of self-driving cars. For instance, D. Dong, X. Li et al. [4] proposed a vision-based method for improving the safety of self-driving cars. They performed road segmentation using SegNet and ENet. Cross entropy loss was applied to the model. ENet produced results faster and with average accuracy but faced problems while upsampling and own sampling. The SegNet method gave better accuracy even when upsampling and downsampling were performed.

P. Prajwal, et al. [5] implemented object detection in self-driving cars using deep learning. They Considered MobileNet as their primary network and built the Caffe MobileNet SSD model by training it with the COCO Dataset. The model produced a mAP value of 72.8%. Chy, Md Kalim Amzad, et al. [6] developed a smart deep learning-based self-driving product. They developed their own datasets using a Udacity self-driving car simulator and performed normalization. They developed a 9-layer neural network model which included 5 consecutive convolutional layers and then the output is flattened and passed into 3 consecutive dense layers with 256,100,10 neurons respectively. Every layer has an ELU activation function with a learning rate of 0.0001 and an Adam optimizer. The model was trained for 20 epochs and produced an accuracy of 84.86%

Similarly, N. Sanil et al. [7] developed a CNN model to train a car to move along a 12 m-long track autonomously. They built a CNN with the first Convolutional Layer with 32 feature extraction followed by 0.2 dropouts followed by 64 feature extraction convolutional layer followed by max pooling, dropout, flattening followed by a dense layer with 128 neurons, and the final dense layer with 4 neurons. Except for the final layer, all other layers use the ReLU activation function, and the final layers use the SoftMax activation function. The model produced an accuracy of 88.6% during the training phase.

## 1.1 Motivation

Self-driving cars are a promising technology that has the potential to revolutionize transportation and improve safety on the road. According to a report by the National Highway Traffic Safety Administration, 94% of crashes are due to human error, and self-driving cars have the potential to significantly reduce the number of accidents and fatalities on the road. However, many challenges must be overcome to make self-driving cars a reality. One important challenge is detecting and avoiding obstacles, such as pedestrians and other vehicles, that may not be included in map data. Other challenges include detecting and classifying traffic signals and taking decisions, accordingly, detecting lanes and predicting the steering angle and acceleration or deceleration based on behavioral cloning.

To address this challenge, many researchers have explored the use of deep learning and computer vision techniques, such as OpenCV, to analyze sensor data from cameras and identify obstacles in real-time. These approaches have achieved impressive results, with some studies reporting accuracy rates of over 90%. However, these approaches can be complex and may require large amounts of labeled data for training. In addition, there is still room for improvement, as current systems may struggle with certain types of obstacles or in challenging lighting conditions.

In this research, we plan to investigate the use of deep learning and OpenCV to detect unexpected obstacles for self-driving cars like vehicles and traffic signs. The ultimate goal is to propose an approach that can achieve good performance on multiclass classification tasks, meaning it can differentiate between different types of obstacles, such as pedestrians, vehicles, and other objects. To do this, I will explore various deep learning architectures and training strategies and evaluate their performance on a range of datasets.

One potential application of this research is in the development of autonomous vehicles for public transportation, such as buses or shuttles. These vehicles must be able to navigate complex environments and avoid obstacles to ensure the safety of passengers and pedestrians. According to the American Public Transportation Association, public transportation saves 37 million metric tons of CO2 emissions annually, and self-driving vehicles have the potential to further reduce emissions and improve the sustainability of transportation systems. By developing more accurate and robust obstacle detection algorithms, we can help make self-driving cars a reality and improve the efficiency and safety of transportation systems.

## 1.2 Objectives

The primary objective of this paper is to explore the application of deep learning and OpenCV techniques in self-driving cars. To achieve the goal, the following specific objectives will be pursued:

- Review the current state of the art in the use of deep learning and computer vision techniques in self-driving cars, including the challenges and limitations of these approaches.

- Propose a novel methodology for integrating deep learning and computer vision techniques into self-driving cars, including the use of image preprocessing and feature extraction techniques to improve the performance of the model.
- Conduct experiments using the proposed approach on a variety of datasets relevant to self-driving cars, including object detection and classification tasks, trajectory prediction, etc., Analyze the results of the experiments and compare the performance of the proposed approach to other state-of-the-art methods.
- Conclude the study by summarizing the key findings and discussing the implications and potential future directions for the application of deep learning and computer vision techniques in self-driving cars.

## 2. Related Work

The following section contains the research works conducted earlier in the field of self-driving cars using deep learning and OpenCV technology.

- Research works that proposed techniques or methods for traffic sign detection:

A Kumar et al. [8] proposed a technique using capsule networks to detect traffic signs. HOG and SIFT techniques were used to preprocess the input image. CapsuleNet model with ReLU activation was proposed which consisted of a group of neurons that encapsulate the outputs into a vector. Comparative analysis was performed between the proposed model and LDA, SVM, RF, and the Capsulenet model producing an accuracy of 96.62%.

W Farag et al. [9] proposed WAF-LeNet which was a CNN based classifier for traffic sign detection. The images were color normalized and Gaussian blur was applied to them. The proposed CNN model consisted of 2 conv layers, 3 FC layers, 2 pooling layers, 3 activation layers, 2 150 dropout layers, and ReLU activation function. The model was tested with the GTSRB dataset with different hyperparameter values. The model reached a testing accuracy of 94.5% with 200 epochs and 128 batch size.

V Swaminathan et al. [10] modified MobileNetv1 architecture for detecting traffic signals. 5 class Belgium traffic sign data was trained for the purpose. An additional SoftMax layer was added to the MobilNet model. The model was deployed into a Raspberry Pi car and used color conversion to grayscale for preprocessing and applied triangle similarity law for distance prediction. The model achieved an accuracy of 83.7%.

T Liang et al. [11] proposed a Sparse RCNN model using ResNeSt backbone architecture for robust traffic sign detection. They integrated the feature pyramid and attention mechanism into the architecture. Further SAA and DTA techniques were used to increase robustness. The model achieved an AP of 99.1% for IOU 75%.

H. Thadeshwar et al. [12] developed a 1/10 scale RC car to portray an automated car. The model used Raspberry Pi to act as a camera, Arduino to control the car, and the laptop as a server. Datasets for the detection of safety signs were COCO datasets. The COCO dataset was trained using the MobileNetv2 model. The model produced an accuracy of 86.5%.

R. Kulkarni et al. [13] used a pre-trained Faster-RCNN-inceptionV2 model for predicting and classifying traffic lights in signals. The model was trained for 120000 iterations for 12 hours and produced a loss of less than 0.01. The datasets were collected from a 16-megapixel camera and resized to 600 x 800 while preprocessing. The model can classify into 5 classes: yellow, red, left right, and straight.

- Research works that proposed techniques or methods for vehicle detection:

C Lin et al. [14] proposed HybridNet a 2-stage vehicle detection algorithm. The first stage involves feature mapping and detecting the bounding boxes and the second stage involves fusing the boxes based on the scores. Inception 4a, 4d, 5b feature maps are concatenated in GoogleNet. NOHEM and NMS techniques were used as post processing steps for the model and data augmentation, Gaussian Blur was used for preprocessing. AP was considered as an evaluation metric and they produced an AP of 74.92 for an IOU of 0.5.

L Chen et al. [15] proposed a lightweight dense neural network for vehicle detection named DLNet. DLNet consists of DenseNet201 architecture as its backbone. In the proposed model consecutive dense blocks are replaced with interlace blocks and group convolutions are employed to reduce BFLOPS. The model produced an AP% of 86% and was 50.1Mb large.

S Mehtab et al [16] proposed a CSPNet based architecture named FlexiNet. It is based on feature pyramid extraction and was trained and tested on the KITTI dataset. The proposed model performs feature fusion using CSPNet blocks. It used SiLU activation functions and batch normalization. Recall and AP with IOU 0.5 were considered as evaluation metrics. The model produced recall of 99.7% for car, 98.3% for pedestrian, 99.3% for cyclist, and an AP of 83.2 with a model size of 54 mb.

Z Yang et al. [17] proposed a YOLOv2 based CNN model for real-time vehicle detection. The model was trained on the KITTI Dataset. The model was modified to use the Priori experience algorithm for feature box extraction rather than the traditional K-means algorithm. Har and HOG features were extracted during preprocessing and Hard Negative Mining technique was used as reinforcement to reduce the errors. Comparative analysis was performed between Faster RCNN, YOLOv2, and the proposed model produced an accuracy of 45% with pedestrians and 61.34% with vehicles in real-time.

- Research works that proposed techniques or methods for lane detection:

Ze Wang et al. [18] proposed a real-time lane detection network named LaneNet. The network works in 2 stages where the first stage classifies the lane edges through pixel-wise edge detection. The second stage of the network takes Inverse Perspective mapping of an image as input and uses Encoder-Decoder architecture for lane localization. Point feature encoder and LSTM decoder were used for robustness. TPR and FPR were considered as evaluation metrics and the models produced a TPR and FPR of 97.9% and 2.75 respectively.

M. khan et al. [19] proposed a state-of-the-art lightweight lane detection approach named LLDnet. Their proposed approach is based on attention-based encoder-decoder architecture. Their model was integrated with channel attention and spatial attention. Their framework was composed of 3 parts-feature extraction phase, convolutional attention block module, and decoder. Comparative analysis was performed between the proposed model and PSPNet, U-net, and FCN based on Dice Coefficient, IoU, AND Dice loss. The proposed model produced a Dice coefficient of 95.46%, IoU of 94.825 and Dice loss of 5.01%.

D. Hau et al. [20] proposed a lightweight UNet based on Depth wise Separable convolutions named DSUNet. This approach was used for path prediction and lane detection. To cluster the detected line segment DBSCAN algorithm was used. The segmentation model was trained and tested using TORCS, LLAMAS, TuSimple image datasets. Accuracy, Precision, Recall, and F1-score were considered as evaluation metrics and DSUNet outperformed UNet by achieving Accuracy, precision, recall, and F1-score of 0.984, 0.768, 0.902, and 0.825 respectively.

Y Huang et al. [21] proposed a spatial temporal based technique for lane detection. For image preprocessing inverse perspective transform and coordinate estimate techniques were used. For training the model a Custom CNN model with 5 layers of CNN followed by 4 layers of FCN were used. Adam optimizer, ReLU, and SoftMax activation function were used. The model produced precision, recall, and F1 scores of 97.067%, 95.389%, and 96.221% respectively for highways and 96.491%, 92.231%, and 94.112% respectively for curvy roads.

Sujeetha, B. Chitrak et al. [22] proposed a method to find lane lines using computer vision. They first converted the image to grayscale then performed Gaussian blur followed by canny edge detection and Hough transform.

- Research works that proposed techniques or methods for behavioral cloning:

MV Smolyakov et al. [23] developed a CNN model to predict the Steering angle in the Udacity Simulator. They collected log data by simulating the car for hours in the simulator. Developed A CNN model with Lambda input layer to normalize the input pixels. The model had 3 conv layers followed by batch normalization dropout and 2 dense layers. Adam optimizer and MSE error functions were used, and the model achieved an accuracy of 78.5%.

A Nguyen et al. [24] proposed FADNet using federated learning techniques for autonomous driving in Udacity simulator. The network trains the distributed data from local machines. The model is based on ResNet8 architecture and uses Global average pooling, Adam optimizer, and RMSE error function. The model had an accuracy of 94.5%.

S Lade et al. [25] proposed a CNN based deep learning model to simulate the car autonomously in Udacity simulator. The images were preprocessed by normalizing the pixels, uv color change, and Gaussian blur. A CNN model was built with 4 conv layers, and 3 dense layers with ELU activation function. The model produced an accuracy of 96.83 and a max error of 0.61.

N Iljaz et al. [26] proposed a hybrid learning model to predict steering angle and direction in Udacity simulator environment. The images were processed using color conversion and normalization. The network consisted of multiple conv, dense, max pooling, and dropout layers followed by time distributed and LSTM layers with ReLU activation, Ada-Delta optimizer, and MSE error function. The proposed model produced an MSE of 0.053.

S. OwaisAli Chishti et al. [27] implemented a self-driving car model using CNN and Q-learning. The dataset was collected by driving the car over the track and images were collected from 3 cameras fitted to it. Supervised learning was performed with the dataset using CNN and reached a training and testing accuracy of 89% and 73% respectively. The reinforcement learning was performed using Deep Q-learning with CNN model for 3 sign boards stop, no left, and traffic light.

A. Bhalla et al. [28] performed a comparative analysis of different neural networks in the simulation of self-driving cars using deep learning. They used Udacity's driving simulator in train mode to collect the datasets for behaviour cloning. 1500 datasets were collected and augmented. They trained the model using 5 models CNN, ResNet50, DenseNet201, Vgg16, and Vgg19. The proposed CNN model consisted of 5 2D-Convolutional layers, One Flatten layer, and four Dense layers. The models were trained for 20 epochs of MSE loss function and Adam optimizer. The CNN produced an MSE loss of 0.029 which was lesser compared to other models.

Ana I. Maqueda et al. [29] developed a CNN model to predict steering angles while driving based on computer vision. They used a dataset of driving sequences recorded with an event-based camera mounted on a car driving around a city. The dataset included both the events from the camera and the corresponding steering angles as ground truth. The authors trained a custom CNN model on this dataset to predict the steering angles from the events. They evaluated the performance of the model using mean absolute error (MAE) as the loss function and reported an MAE of 0.21 radians on the test set.

- A few recent works on autonomous driving:

The research conducted by Jian et al.[30], identified the issues when path-planning algorithms were applied to traditional planning methods and the reasons for their occurrence. It was observed that the traditional planning methods affect a vehicle's traveling path, comfort, and velocity. In response to these problems, a framework was proposed that includes three layers active path planning layer, the velocity planning layer, and the real-time scheduling layer. The proposed method was tested on several real-life traffic scenarios to prove its performance.

The performance of the single and multi-modal models using a simulation environment for autonomous driving was analyzed by Haris and Glowacz [31], in their work. The information obtained from the simulation environment is post-processed and is used to train a depth estimation model. The results of the experiments conducted in their research show that the multi-modal model performs well in all scenarios when compared to single models. Their work also mentions the major limitations of the proposed work and the scope for future research.

The study conducted by Cui, Han, and Liu [32], identified the limitations of the existing models, and their infeasibility for online applications. To resolve the issue a model was proposed which included three layers, the first layer refines the object representations and predictions from the prior stage, and the second layer fuses the local information from the neighbouring frames and global semantics from the current frame to prevent feature degradation. The third layer is based on the refinement results to eliminate redundancy and improve performance to stop aggregation early.

A method based on deep learning was proposed by Liu et al.[33], for video object detection. The proposed model was able to obtain a better trade-off between accuracy and speed. The core concept focuses on utilizing the temporal coherence of video features while accounting for motion patterns captured through optical flow. To enhance detection accuracy, the framework combines calibrated features at both the pixel and instance levels across multiple frames, improving robustness against appearance changes. This aggregation and calibration are performed adaptively and efficiently using an integrated optical flow network. Furthermore, the proposed method features an end-to-end architecture, which significantly boosts training and inference efficiency compared to traditional multi-stage approaches in video object detection.

A novel method was proposed by Liu et al. [34], for visual localization. The proposed methods concentrate on three views front, left, and right compared to conventional methods which improve the effectiveness of the prediction of location. In their work, three visual localization networks were modified for vehicular operation. The proposed model was evaluated under different challenging conditions to prove its performance.

A dataset was developed by Cui et al.[35], which is efficiently labeled, represents the diversity of data, and temporal representations which proved to be better when compared to the previous attempts and therefore helped to increase the accuracy and efficiency of training. Various experiments were carried out to test and prove the performance of the dataset developed. An annotator was also proposed which can produce annotations of good quality.

## 3. Datasets and Methods

In this section, we will discuss the datasets taken for training different models. We will discuss in detail the different pre-processing techniques applied to the datasets, the architecture of different models taken for training the datasets, etc.,

### 3.1 Dataset

German Traffic Sign Recognition Benchmark (GTSRB) datasets were collected from the Kaggle repository for this research purpose. Fig.1 shows the sample images from the dataset. The dataset consisted of a total of 73139 images belonging to 43 classes. The images in the datasets were pre-processed and cropped exactly to fit the traffic sign. The 43 classes in the dataset represent the 43 different traffic signs. Each image in a particular class is auto encoded to a number representing the class. Included images like turn right, left, speed limits, etc., The data was augmented using rescaling, flipping, shearing, and zooming techniques.

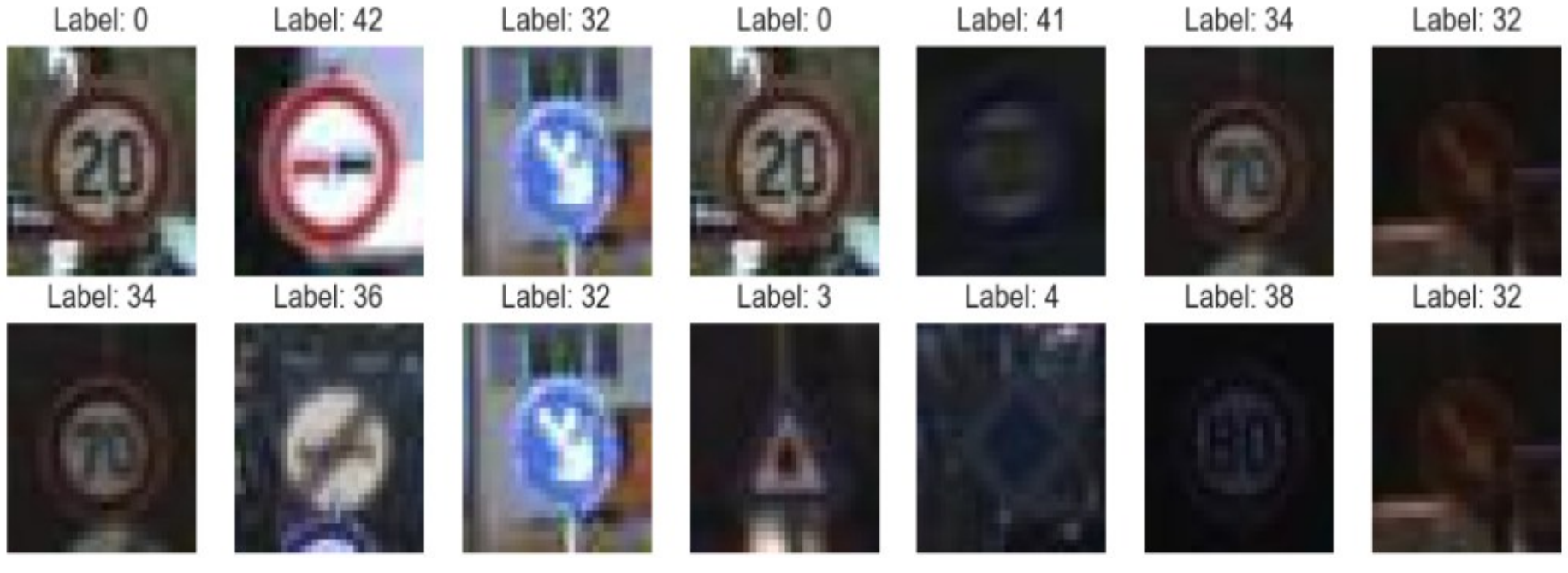

Fig. 1 German Traffic Sign Recognition Benchmark (GTSRB) Dataset

Lane Datasets were collected from the Kaggle repository for this research purpose. The datasets consisted of 2 training datasets where one dataset consisted of real-life images of the roads and lanes and the other dataset consisted of masked images of the respective roads after undergoing image segmentation. The dataset consisted of a total of 287 images which belonged to 3 classes. The 3 classes include normal roads in residential areas, highways and lanes, and the respective masked images. The dataset was divided into train, validation, and test as 231,28 and 30 images respectively. Fig. 2 and Fig. 3 show the original and corresponding Image of Segmentation Mask (right) of a Lane, and the Original (left) and corresponding Image of Segmentation Mask (right) of a Road respectively. The dataset also consists of images with yellow end lines and white yellow lines. Fig. 4 and Fig. 5 show a highway with a yellow line and a highway with an end line respectively.

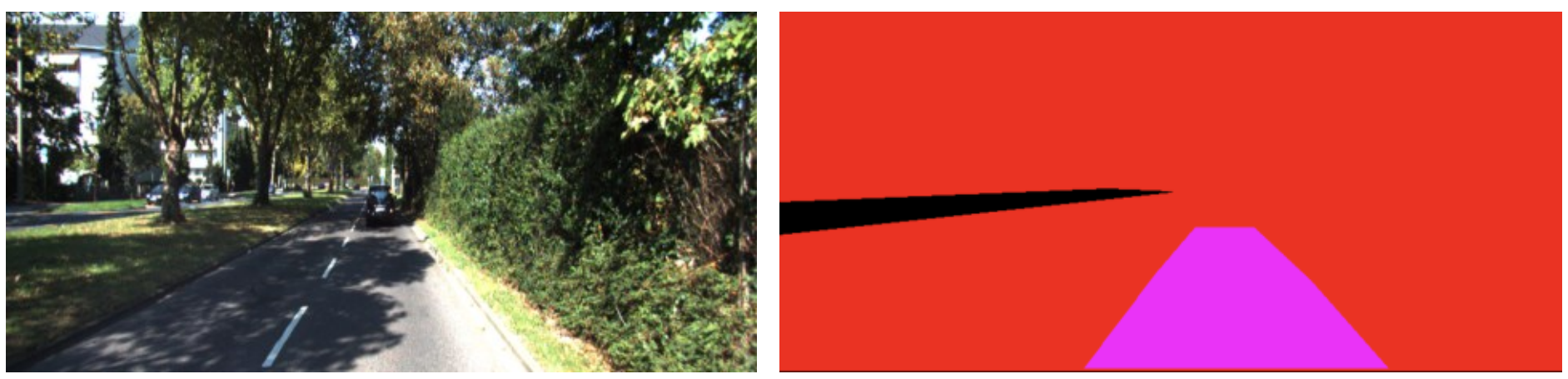

Fig. 2 Original (left) and Corresponding Image of Segmentation Mask (right) of a Lane

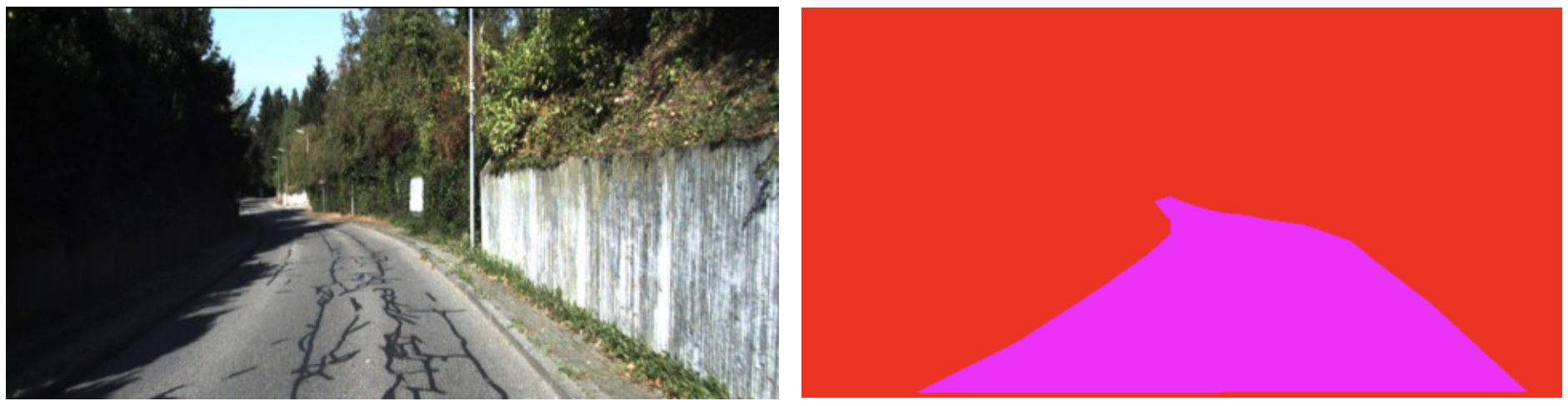

Fig. 3 Original (left) and Corresponding Image of Segmentation Mask (right) of a Road

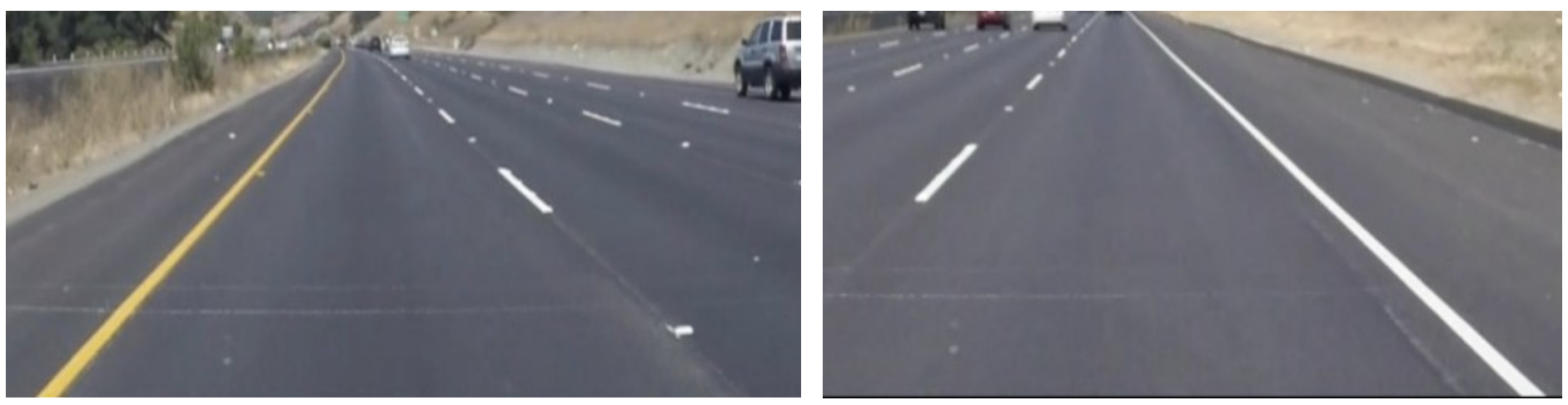

Fig. 4 Highway with Yellow Line Fig. 5 Highway with White Line

Vehicle Datasets were collected from the Kaggle repository for this research purpose. The datasets consisted of a total of a total of 17760 images belonging to 2 classes vehicle and non-vehicles. The vehicle class included cars, trucks, bicycle, tram, motorcycle, and other road vehicles. The sample images from the dataset are shown in Fig. 6. The dataset was further split into train and test datasets in the ratio 80:20 and the final train set consisted of 14208 images and the test set consisted of 3552 images. The datasets were further preprocessed using the following techniques: shear, zoom, vertical and horizontal split, height, and width shift. It was finally resized to 75x75 since ResNet can take input of the shape 75x75. It was further reshaped into a 32x32 image size since Custom CNN could take input shape as 32x32.

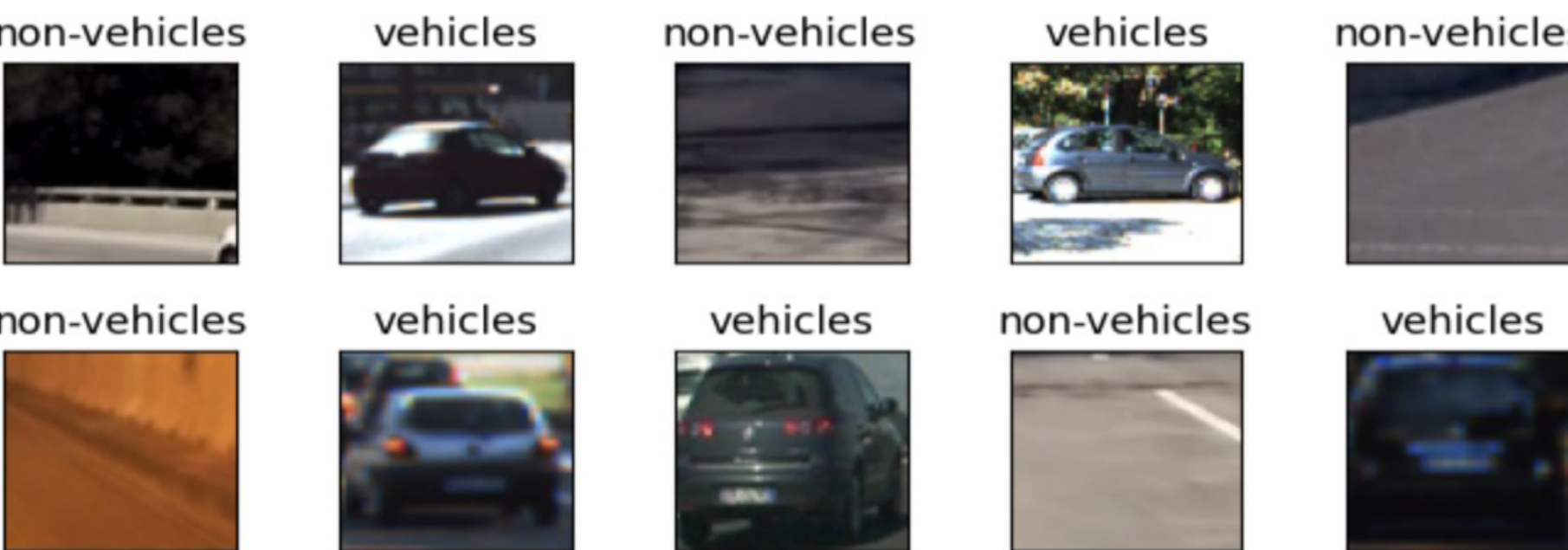

Fig. 6 Vehicle Dataset

Datasets for Behavioural Cloning were collected from the Kaggle repository for this research purpose. The datasets can alternatively be acquired by driving the car manually in Udacity's self-driving car simulator and recording the car movement which collects images and steering angle, throttle, reverse, and speed values. The datasets taken for the research purpose consisted of images collected from 3 cameras of the car: center, left, and right. Additionally, the dataset consisted of steering, throttle, reverse, and speed values at respective points of driving.

The image datasets were merged with corresponding steering values, throttle values, reverse values, and speed values through Panda's library. The dataset after merging consisted of 7 parameters and 3403 sample cases. The datasets were preprocessed by different techniques like cropping, RGB to UV color transformation, applying Gaussian blur, flipping, and resizing the images to 200 x 66. The datasets were further split into the ratio 80:20. The training set consisted of 5596 samples. Fig. 7 and Fig. 8 show the dataset collected from the Udacity self-driving car simulator, and the dataset after merging steering, and throttle values with images respectively.

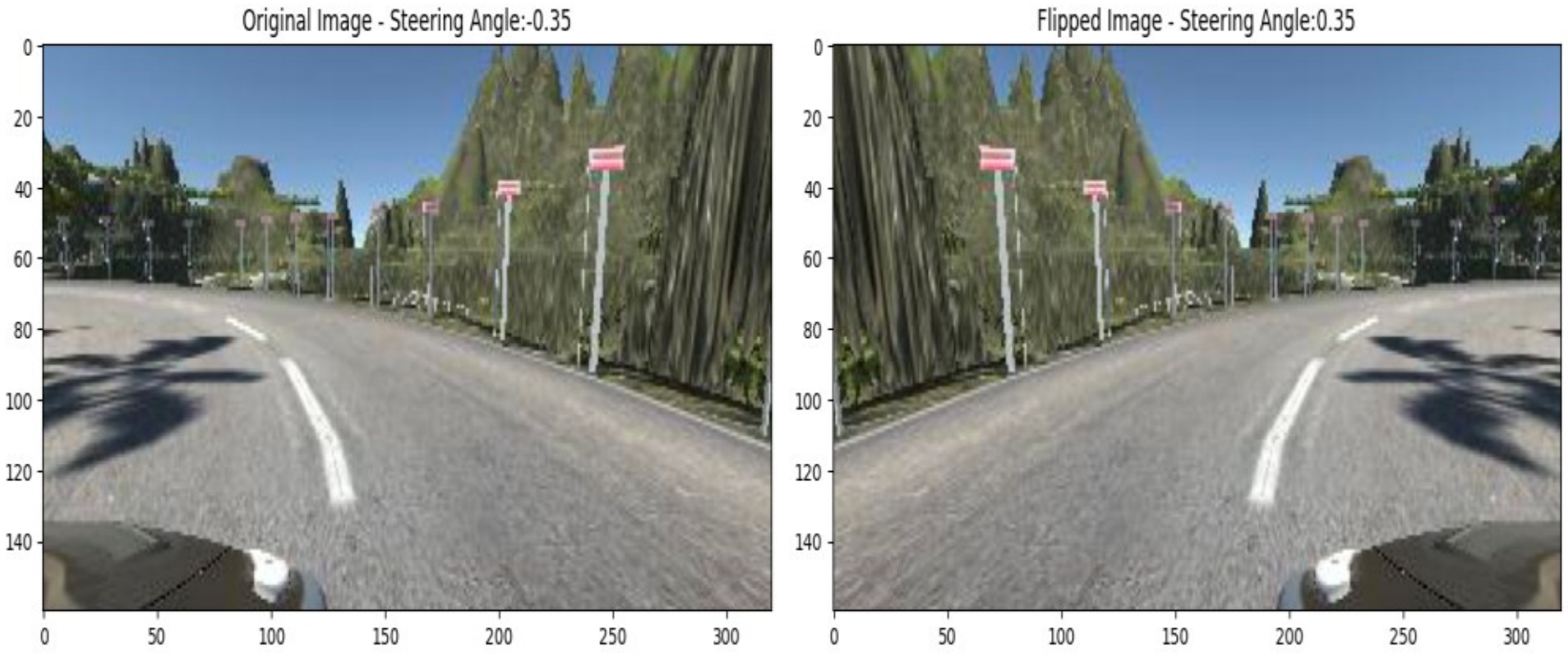

Fig. 7 Dataset Collected from Udacity Self-driving Car Simulator

| center | left | right | steering | throttle | reverse | speed |
|---|---|---|---|---|---|---|
| center_2022_04_10_12_44_27_913.jpg | left_2022_04_10_12_44_27_913.jpg | right_2022_04_10_12_44_27_913.jpg | 0.00 | 1.0 | 0 | 21.69488 |
| center_2022_04_10_12_44_27_983.jpg | left_2022_04_10_12_44_27_983.jpg | right_2022_04_10_12_44_27_983.jpg | 0.00 | 1.0 | 0 | 22.50011 |
| center_2022_04_10_12_44_28_052.jpg | left_2022_04_10_12_44_28_052.jpg | right_2022_04_10_12_44_28_052.jpg | 0.00 | 1.0 | 0 | 23.11461 |
| center_2022_04_10_12_44_28_121.jpg | left_2022_04_10_12_44_28_121.jpg | right_2022_04_10_12_44_28_121.jpg | 0.00 | 1.0 | 0 | 23.89061 |
| center_2022_04_10_12_44_28_191.jpg | left_2022_04_10_12_44_28_191.jpg | right_2022_04_10_12_44_28_191.jpg | 0.05 | 1.0 | 0 | 24.46815 |

Fig. 8 Dataset After Merging Steering, Throttle Values with Images

## 3.2 Data Pre-processing

### I. Color Segmentation

In color segmentation, each pixel is represented by a single intensity value ranging from 0 (black) to 255 (white). Grayscale conversion is commonly used in lane detection because it simplifies the image processing task. Fig. 9 shows the output of Grayscale conversion. Lane detection algorithms often rely on edge detection techniques to identify the lane markings, and grayscale conversion improves the accuracy of edge detection. It also reduces the computational load of the algorithm as grayscale images have only one channel instead of three channels in a color image. This makes the algorithm faster and more efficient.

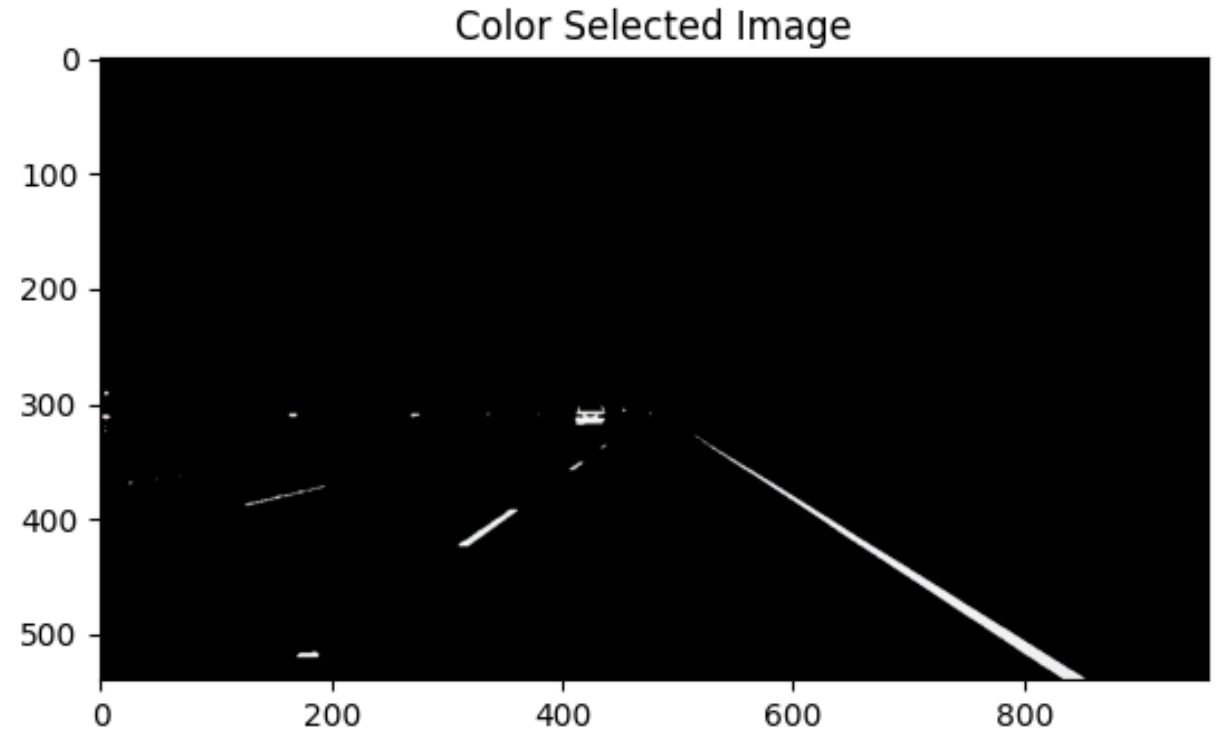

Fig. 9 Output of Grayscale Conversion

### II. Gaussian Blur

The Gaussian blur filter works by taking the weighted average of each pixel in an image with its neighboring pixels, with the weights determined by the Gaussian distribution. This blurs the sharp edges and details in an image, resulting in a smoother and less noisy image. Fig. 10 shows the output of Gaussian Blur. The 2D Gaussian filter can be defined mathematically as:

$$G\,(x,\,y) = \frac{1}{2\Pi\sigma^2}\, e^{-\frac{x^2+y^2}{2\sigma^2}} \qquad (1)$$

where x and y are the pixel positions, sigma is the standard deviation of the Gaussian distribution, and exp is the exponential function.

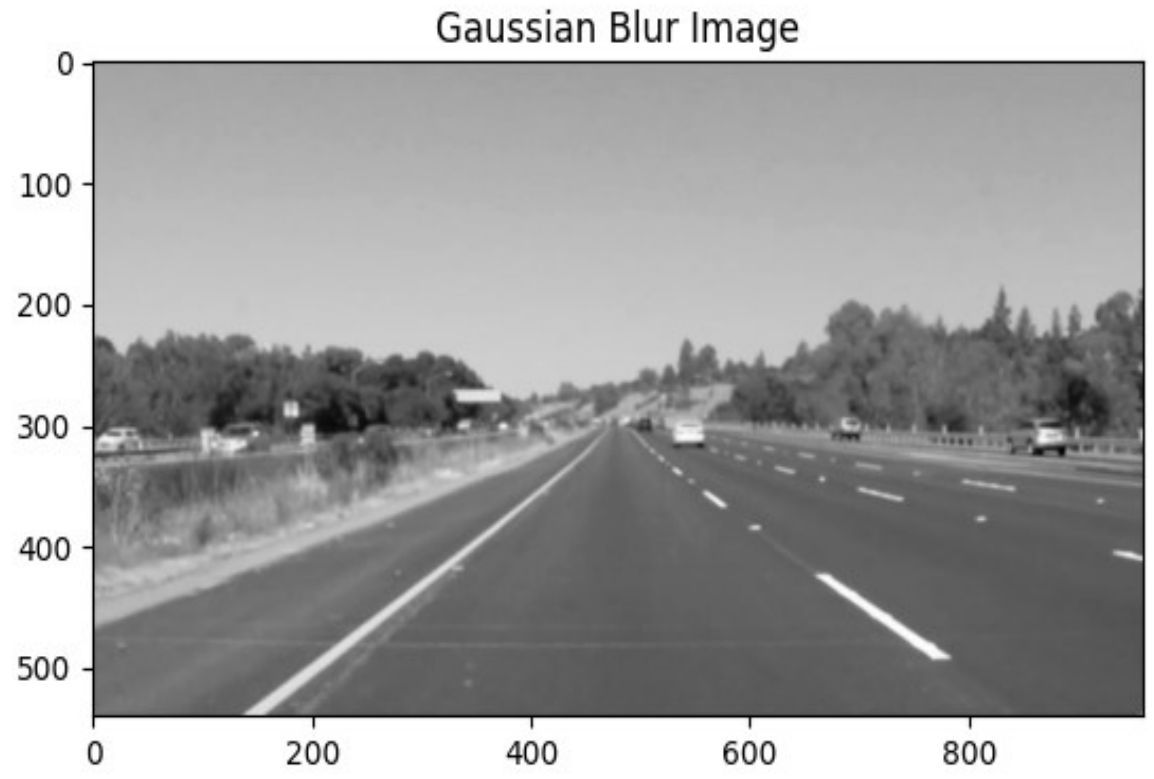

Fig. 10 Output of Gaussian Blur

## III. Canny Edge Detection

After applying Gaussian blur the edges are located by calculating the Sobel filter. Using the Non-maximum suppression algorithm, the pixel values are assigned 0 if it's not a local maximum. The final step includes thresholding and combining the weak edges with the strong edges to make a continuous edge. This is very useful in case there are yellow marking lanes in the road that grayscale can't detect. The below Fig. 11 shows output of Canny Edge Detection.

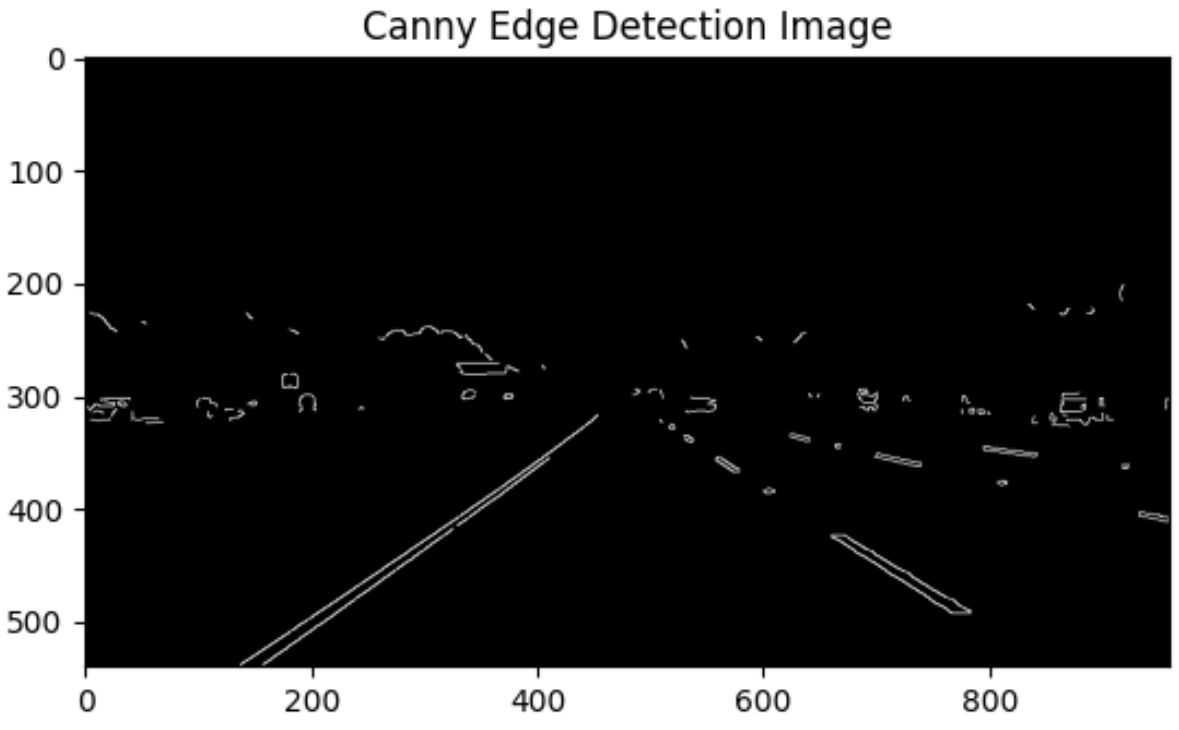

Fig. 11 Output of Canny Edge Detection

## IV. ROI Masking

We define the area in which we want to apply the filters, mostly the road area is the region of interest. We mask it with a polygon like a triangle or trapezium. Fig. 12 shows the output of ROI Masking.

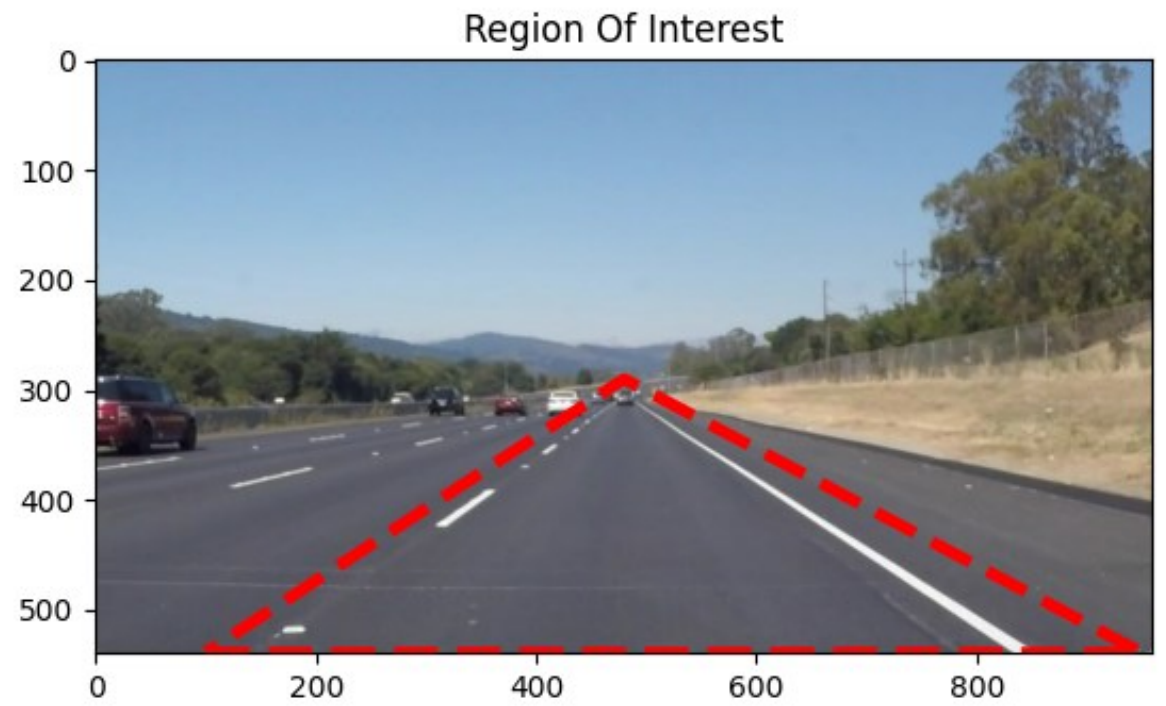

Fig. 12 Output of ROI Masking

## V. Hough Transform

Hough transform can detect lines and curves that may not be apparent in the original image due to noise or other factors. In lane detection, the Hough transform is applied to the output of the canny edge detection algorithm to detect lines that represent the lane boundaries. The output of the Hough Transform is shown in Fig. 13.

In the standard Hough transform for line detection, the equation of a line is represented in polar coordinates as:

$$r = x \cos(\theta) + y \sin(\theta) \quad (2)$$

Where (x, y) are the coordinates of a point on the line, theta is the angle between the line and the x-axis, and r is the distance from the origin to the line.

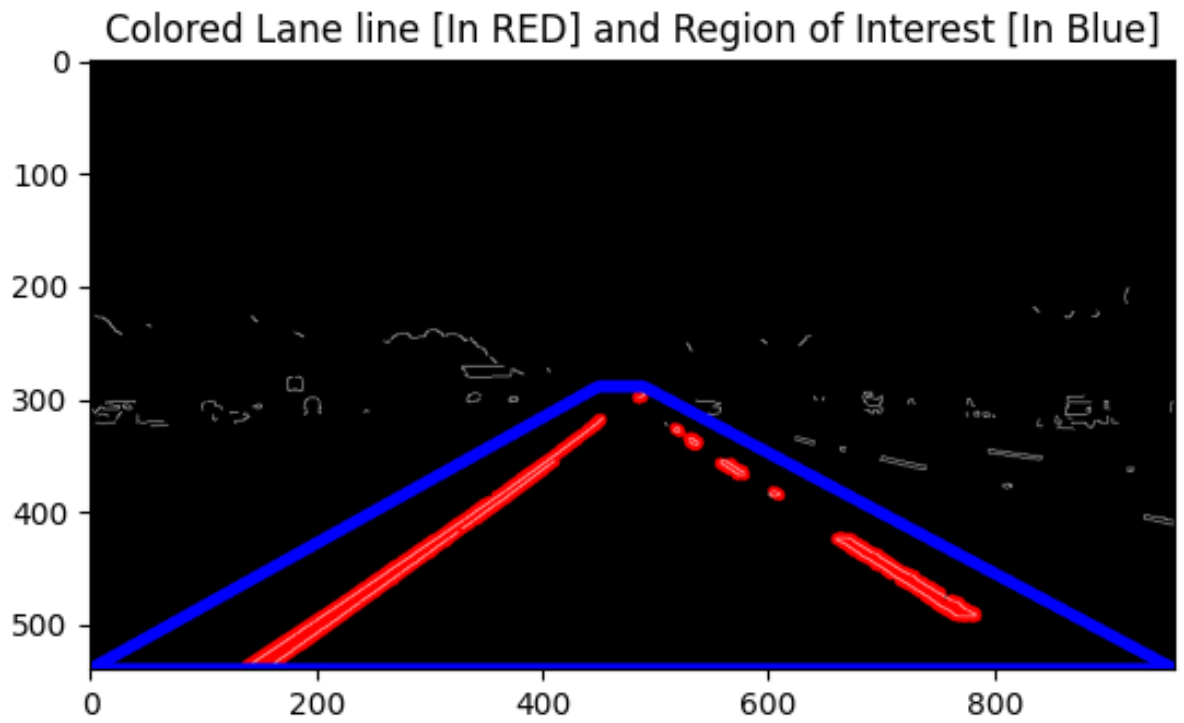

Fig. 13 Output of Hough Transform

## VI. Auto-Encoder

An autoencoder is a type of artificial neural network that is trained to encode and decode data. It consists of two parts: an encoder and a decoder. The encoder takes an input and converts it into a lower-dimensional representation, while the decoder takes this representation and attempts to reconstruct the original input.

The autoencoder is trained using unsupervised learning, which means that it does not require labeled data. The network is trained to minimize the difference between the input and output, which encourages the encoder to learn a compressed representation of the data that captures the most important features.

In the study, we are considering a VGG16 backbone and upsampling the pooling layers of block 3,4,5 and concatenating them with a 2x2 filter and the decoder layer reconstructs the input through a convolutional 2D layer.

## 3.3 Transfer Learning and Fine Tuning

Transfer learning is a machine learning technique where a pre-trained model is used as a starting point for solving a new task. Instead of building a new model from scratch, transfer learning allows us to leverage this pre-trained model's knowledge to solve a similar or related problem. We can use the pre-trained model's weights as a starting point and fine-tune them using our dataset to train a new model that is tailored to our specific task. Some Transfer Learning models taken for research in this work are ResNet50, VGG16, MobileNet, InceptionV3, Xception, etc., The models consisted of replicative layers of convolution, activation, and pooling layers. The last layer was removed from the models and modified according to our research.

I.Max Pooling:

Max pooling is a technique used in convolutional neural networks (CNNs) to reduce the spatial size of the feature maps generated by the convolutional layers. This results in a new, smaller feature map that retains the most important information from the original feature map while reducing its dimensionality.

MaxPooling is defined mathematically as:

$$f(x) = \max(x_i) \tag{3}$$

Where i is the index of the window. Where $x_i$ is the window of the feature map where the max pooling operation is applied.

II.Dropout

Dropout is a regularization technique used in neural networks, to prevent overfitting. It randomly drops out a certain percentage of the neurons in the network during training. This forces the network to learn multiple independent representations of the input data, which helps to reduce overfitting and improve the generalization of the model.

III. Batch Normalization

Batch normalization is a technique that works by normalizing the inputs to a layer by adjusting and scaling them to have a mean of zero and a variance of one. This is done for each batch of inputs fed to the layer during training. It helps in reducing the internal covariate shift problem and accelerates the training process.

IV. Activation Functions

Sigmoid:

The sigmoid function is a widely used activation function that maps any input value to a range between 0 and 1. The function has a characteristic S-shaped curve and is defined as:

$$f(x) = \frac{1}{(1+e^{-x})} \tag{4}$$

ReLU (Rectified Linear Unit):

ReLU is another widely used activation function. It maps any input value below 0 to 0 and any positive value to the value itself. The function is defined as:

$$f(x) = \max(0, x) \tag{5}$$

ELU:

ELU is similar to ReLU, but negative inputs have different outputs instead of 0.

$$f(x) = \begin{cases} x & x \geq 0 \\ \alpha(e^{x} - 1) & x < 0 \end{cases} \tag{6}$$

SoftMax:

SoftMax is often used as the output layer activation function for multi-class classification problems. It maps the input values to a probability distribution over the classes. The function is defined as:

$$f(x) = \frac{e^{x}}{\sum e^{x}} \tag{7}$$

## V. Loss Functions

Mean Squared Error (MSE):

MSE is a common loss function used for regression problems. It measures the average squared difference between the predicted and actual values. The equation for MSE is:

$$MSE = \frac{1}{n} \sum_{i=1}^{n} ( Y_i - \hat{Y}_i )^2 \tag{8}$$

where $Y_i$ is the observed values and $\hat{Y}_i$ is the predicted value and n is the no. of data points.

Root Mean Squared Error (RMSE):

RMSE is similar to MSE but takes the square root of the average squared difference. The equation for RMSE is:

$$RMSE = \sqrt{\frac{\sum_{i=1}^{n} ( Y_i - \hat{Y}_i )^2}{n}} \tag{9}$$

Where $Y_i$ is the observed values and $\hat{Y}_i$ is the predicted value and n is the no. of data points.

Categorical Cross-Entropy:

Categorical cross-entropy is a loss function commonly used in multi-class classification problems. It measures the dissimilarity between the predicted probability distribution and the true probability distribution. The equation for categorical cross-entropy is:

$$L(\hat{y},y) = -\frac{1}{N} \sum_{i}^{N} [y_i \log \hat{y}_i + (1 - y_i) \log(1 - \hat{y}_i)] \tag{10}$$

Where $Y_i$ is the observed values and $\hat{Y}_i$ is the predicted value and n is the no. of data points.

Cross-Entropy:

Cross-entropy is a generalization of categorical cross-entropy that can be used for both binary and multi-class classification problems. The equation for cross-entropy is:

$$H(P^* \mid P) = -\sum_i P^*(i) \log P(i) \quad (11)$$

Where $P^*$ and $P$ are the probabilities of event A and event B happening.

### Binary Cross-Entropy:

Binary cross-entropy is a special case of cross-entropy that is used for binary classification problems. The equation for binary cross-entropy is:

$$H_p(q) = -\frac{1}{N} \sum_{i=1}^{N} y_i \,.\log(p(y_i)) + (1 - y_i) \,.\, \log(1 - p(y_i)) \quad (12)$$

## VI. Optimizers

### a. SGD (Stochastic Gradient Descent):

SGD is a basic optimizer that updates the weights based on the gradient of the loss function with respect to the weights. It uses a fixed learning rate for all weight updates, which can lead to slow convergence.

The update rule for SGD is as follows:

$$\theta_j = \theta_j - \alpha \frac{\partial}{\partial \theta_j} J(\theta) \quad (13)$$

were,

$\theta_j$ : gradient at time step j

$\alpha$ : learning rate.

### b. RMSprop (Root Mean Square Propagation):

RMSprop is another adaptive learning rate optimizer that scales the learning rate for each weight based on the average of the magnitudes of recent gradients for that weight. It keeps a moving average of the squared gradient for each weight and divides the learning rate by the root of this average.

The update rule for RMSprop is as follows:

$$\nu_t = \beta\, \nu_{t-1} + (1- \beta) * g_t^2 \quad (14)$$

$$\omega_{new} = \omega_{old} - \frac{\eta}{\sqrt{\upsilon_{t} + \varepsilon}} * g_t \quad (15)$$

### c.Adam (Adaptive Moment Estimation):

Adam is an adaptive learning rate optimizer, which means it adapts the learning rate during training. It combines the advantages of two other stochastic gradient descent (SGD) extensions - AdaGrad and RMSprop. Adam keeps a moving estimate of the first and second moments of the gradient and uses them to adjust the learning rate. It also includes bias-correction to correct the bias that occurs in the estimation of the first and second moments.

The update rule for Adam is as follows:

$$m_t = \beta_1 * m_{t-1} + (1 - \beta_1) * \nabla\omega_t \quad (16)$$

$$\nu_t = \beta_2 * \nu_{t-1} + (1 - \beta_2) * \nabla\omega_t \quad (17)$$

$$\widehat{m_t} = \frac{m_t}{1 - {\beta_1}^t} \quad (18)$$

$$\widehat{\upsilon_t} = \frac{\upsilon_t}{1 - {\beta_2}^t} \quad (19)$$

$$\omega_{t+1} = \omega_t - \frac{\eta}{\sqrt{\widehat{\upsilon_t} + \varepsilon}} * \widehat{m_t} \quad (20)$$

## 4. Proposed System Architecture and Workflow

### 4.1 Traffic Sign Detection

- ResNet50

The images in the datasets were reshaped to the size 32x32. The dataset was split up into train and test in the ratio 80:20 respectively. The model was trained with the help of Transfer learning using Reset50. Global average pooling was applied to the model and a dropout of 0.5 was applied. The output after dropout was passed into a Dense layer with a SoftMax activation function. The dense layer is the final layer of the model and makes the prediction. The model was trained for 25 epochs with a batch size of 32 and a categorical loss function. The metrics used for evaluation were accuracy and loss. The model architecture modified is shown in Fig. 14.

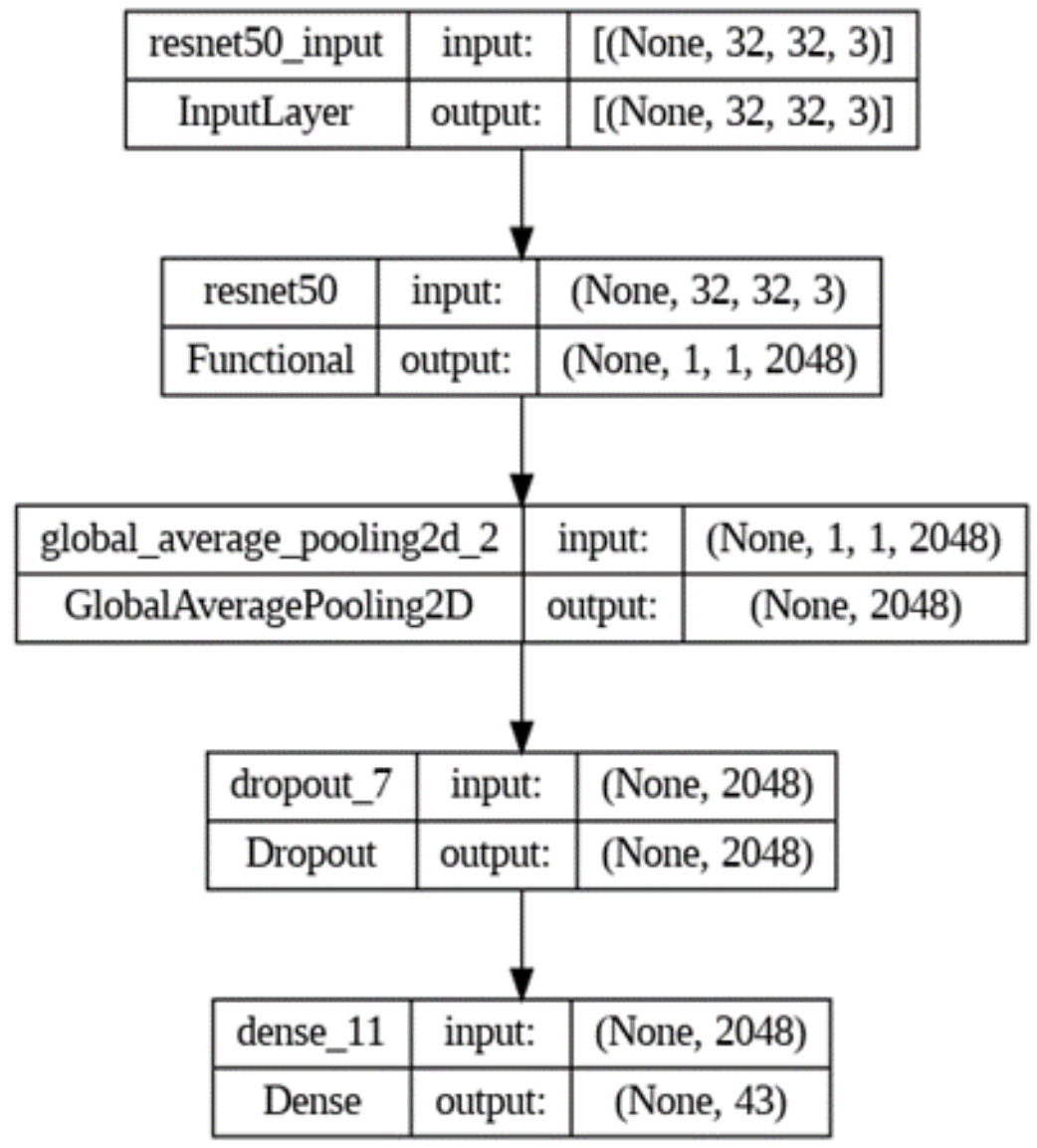

Fig. 14 Modified ResNet50 Model Architecture

- Proposed CNN Model:

The images in the datasets were reshaped to the size 32x32. The dataset was split up into train and test in the ratio 80:20 respectively. The model was trained with the help of a Custom CNN model. The model was created using 3 convolutional layer blocks with the first block consisting of a convolutional 2D layer with 64 filters, kernel size of 3x3, ReLU activation function followed by max-pooling with 2x2 stride length followed by batch normalization and 30% dropout. This block was followed by 2 similar convolutional layer blocks with the same configurations except for the number of filters which were 128 and 512 respectively. The proposed model for traffic sign detection is shown in Fig. 15.

The model was flattened to convert it from an array format into a vector format. The output after Flattening was passed into 2 Dense layers with 4000 neurons each and a ReLU activation function. The output is then passed onto another Dense layer with 1000 neurons and a ReLU activation function. The output was then passed onto the final dense layer with 43 neurons and a SoftMax activation function. The final dense layer of the model makes the prediction. The model was trained for 20 epochs with a batch size of 64 and categorical cross-entropy loss function and Adam optimizer. The metrics used for evaluation were accuracy and loss. The output prediction after ResNet50 based proposed model is as shown in Fig. 16.

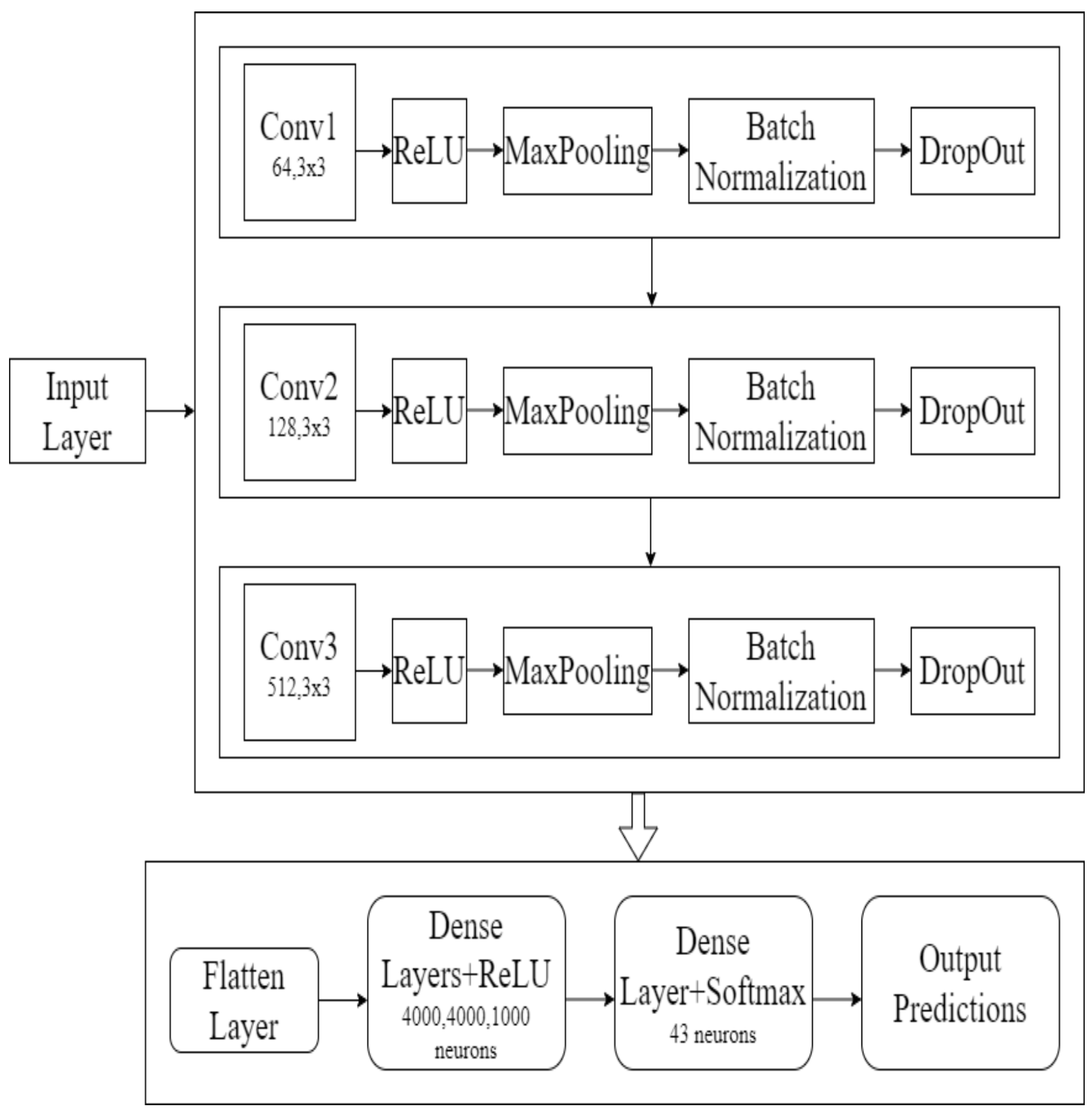

Fig. 15 Proposed CNN Model Architecture

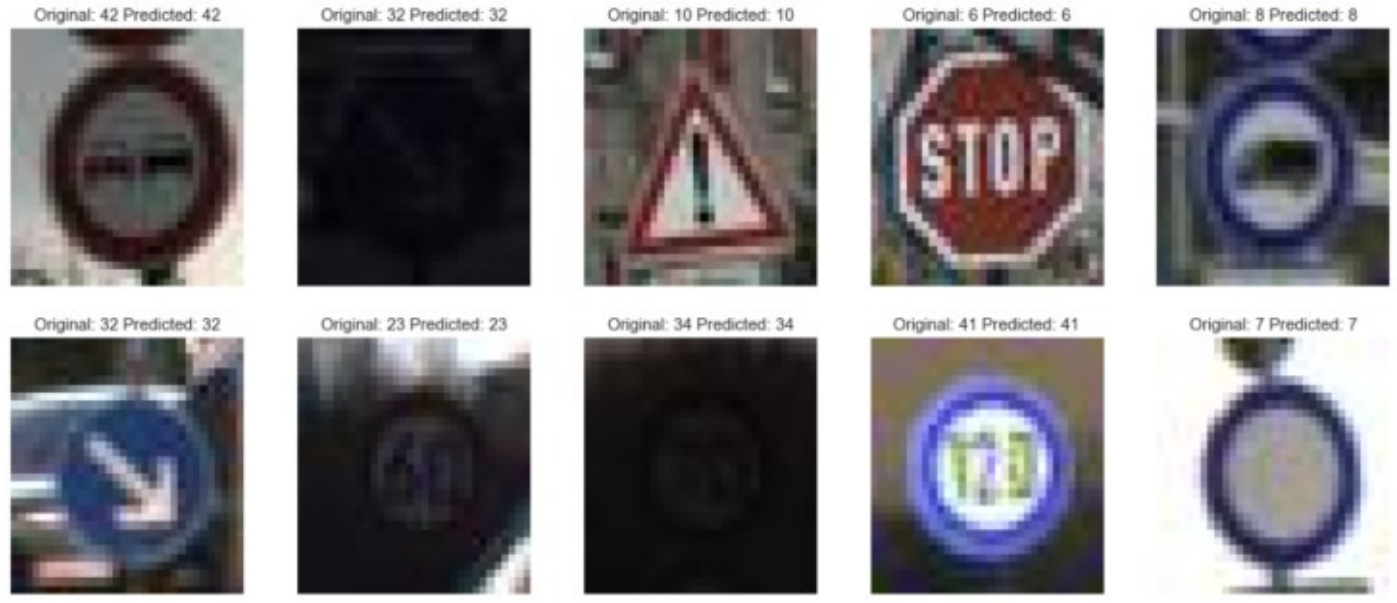

Fig. 16 Output Prediction of ResNet50 Model After Training

## 4.2 Lane Detection:

- Image Segmentation using FCNN with VGG16 Backbone:

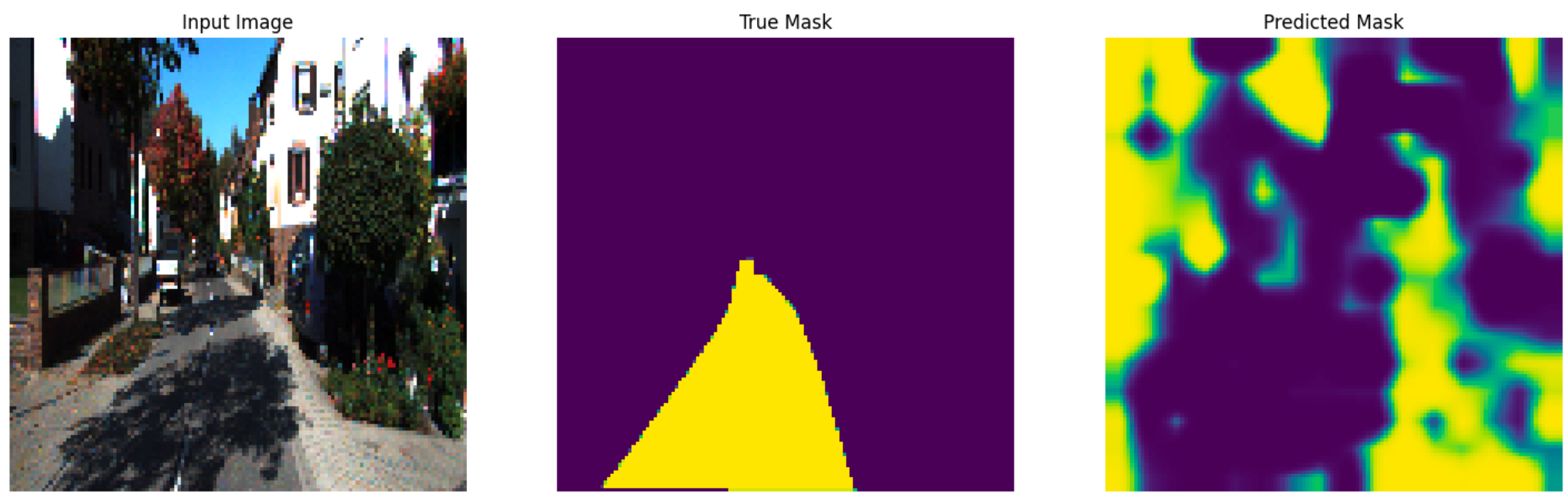

Fig. 17 Output Prediction of VGG16 Model After Training

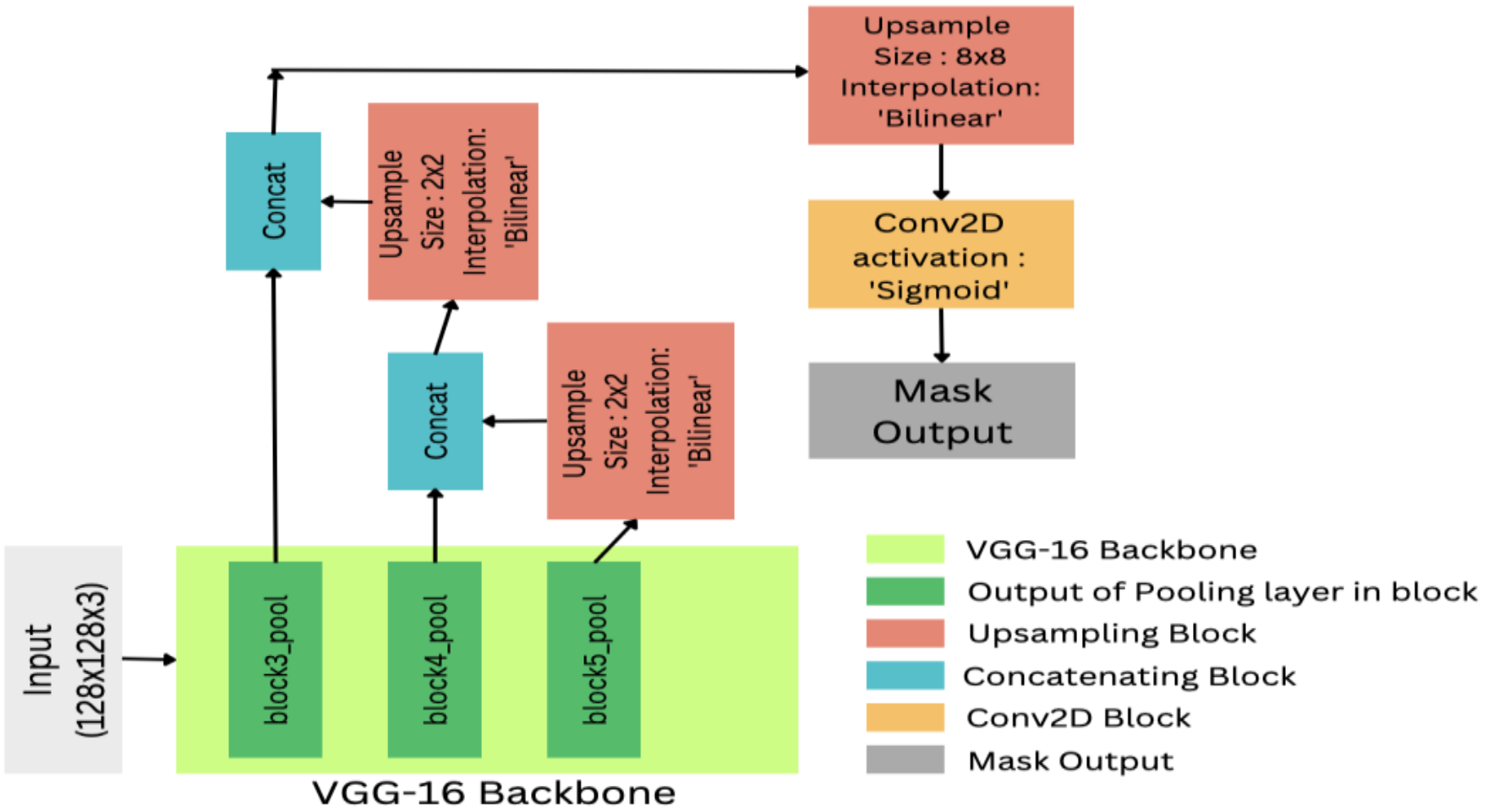

Fig. 18 Proposed Model Workflow Diagram

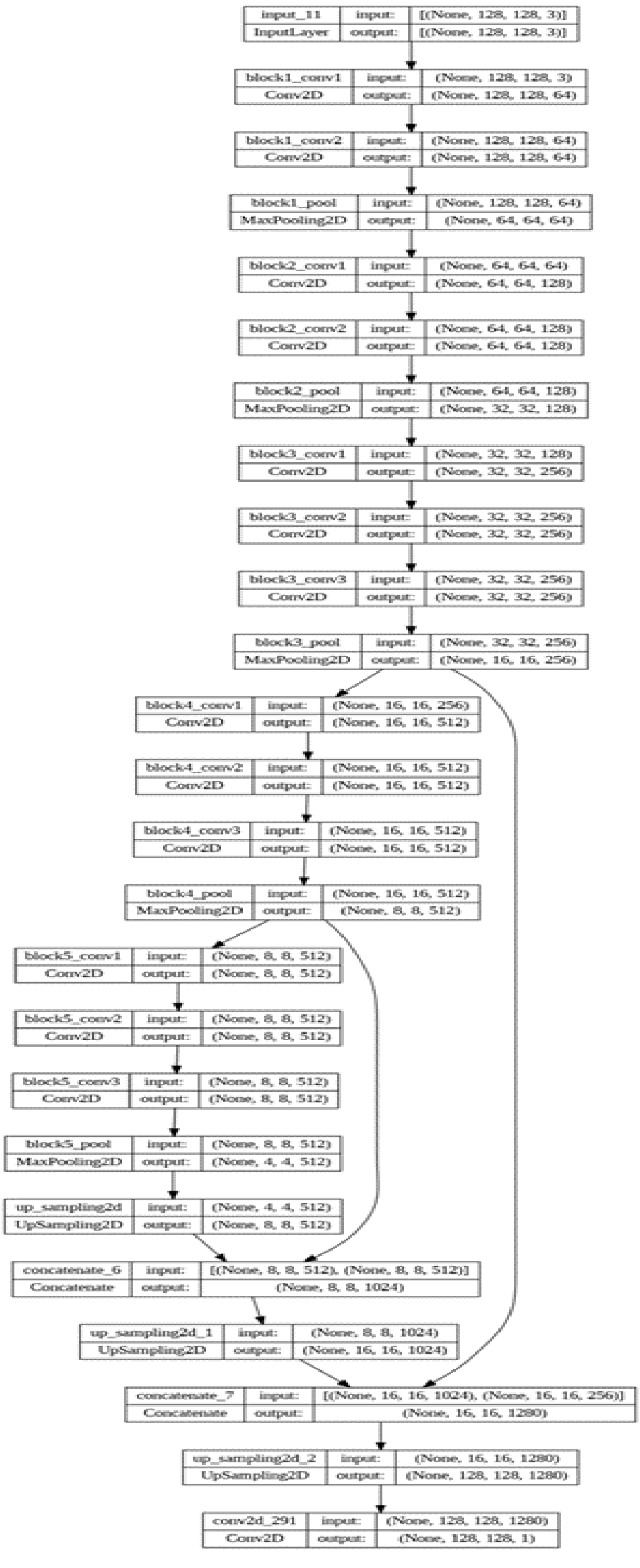

Fig. 19 Proposed FCNN Model with VGG16 Backbone Architecture

The datasets consisted of both real-life images and respective masked images. The real-life images and masked images were fused together using TensorFlow functions. The image input shape was 128x128. The model was trained with VGG16 as the backbone model architecture. The weights were downloaded from ImageNet. The output of pooling layers of blocks 3,4,5 of the VGG16 backbone was extracted using Encoders. The output of pooling layer 5 was up sampled by a factor of 2 and concatenated with the output of pooling layer 4. The concatenated output was again up sampled and concatenated with pooling layer 3. The concatenated output was finally up sampled with a factor of 8.

All the upsampling was performed through bilinear interpolation in the Decoder layer. The output of up-sampling was passed onto a conv2d block with sigmoid activation which produced a masked output. The model was trained for 20 epochs with a batch size of 32. The model was trained with a Binary cross entropy activation function and three different optimizers: Adam optimizer, RMSprop, and SGD to perform comparative analysis. The metric chosen to evaluate the model performance was m_iou (Mean Intersection of Union). Fig. 17 and Fig. 18 show the output prediction of the model after training and the proposed model workflow respectively. Fig. 19 shows the complete layer wise backbone architecture of the Proposed FCNN Model with a VGG16 Backbone.

- OpenCV Transforms:

  The first part of the research included a color selection technique by setting threshold values for red, green, and blue colors in the image. Fig. 20 shows the prediction of lanes using color selection. The second part of the research included creating a set boundary or polygon using the coordinates of the image and extracting the colors inside the specified region. The third part of the research included using the canny edge detection method. Fig. 22 shows the output prediction of lanes after Canny edge Detection. The image was first converted into grayscale and later gaussian blur was applied to remove noise from the data. Finally, by setting the lower and higher threshold the model was able to predict the edge lines of the image. The final stage of lane detection included creating a pipeline that defines the flow in which the images must undergo transforms before making predictions. The complete process flow diagram of lane detection is shown in Fig. 21. Fig. 23 shows the output prediction of lanes after passing through a pipeline of transforms.

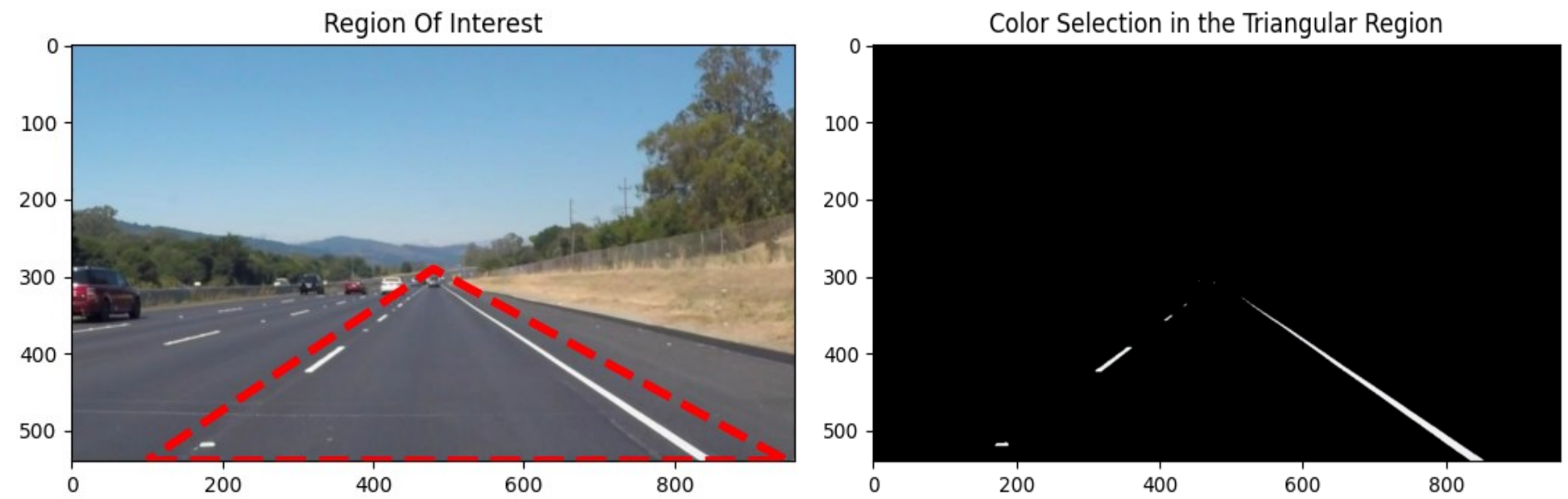

Fig. 20 Prediction of Lanes Using Color Selection

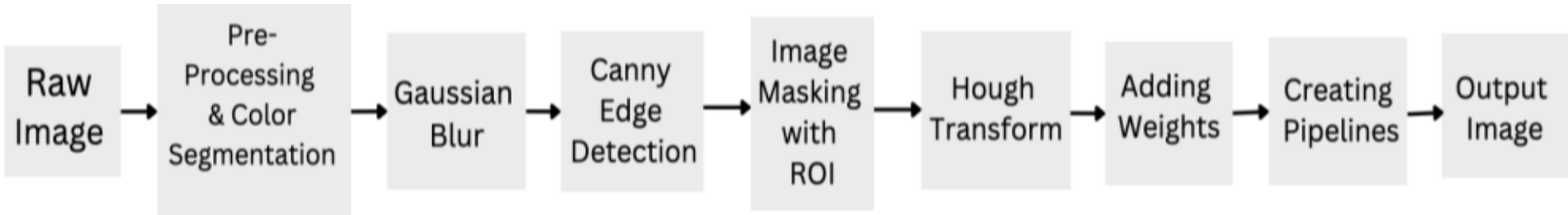

Fig. 21 Process Flow Diagram of Lane Detection

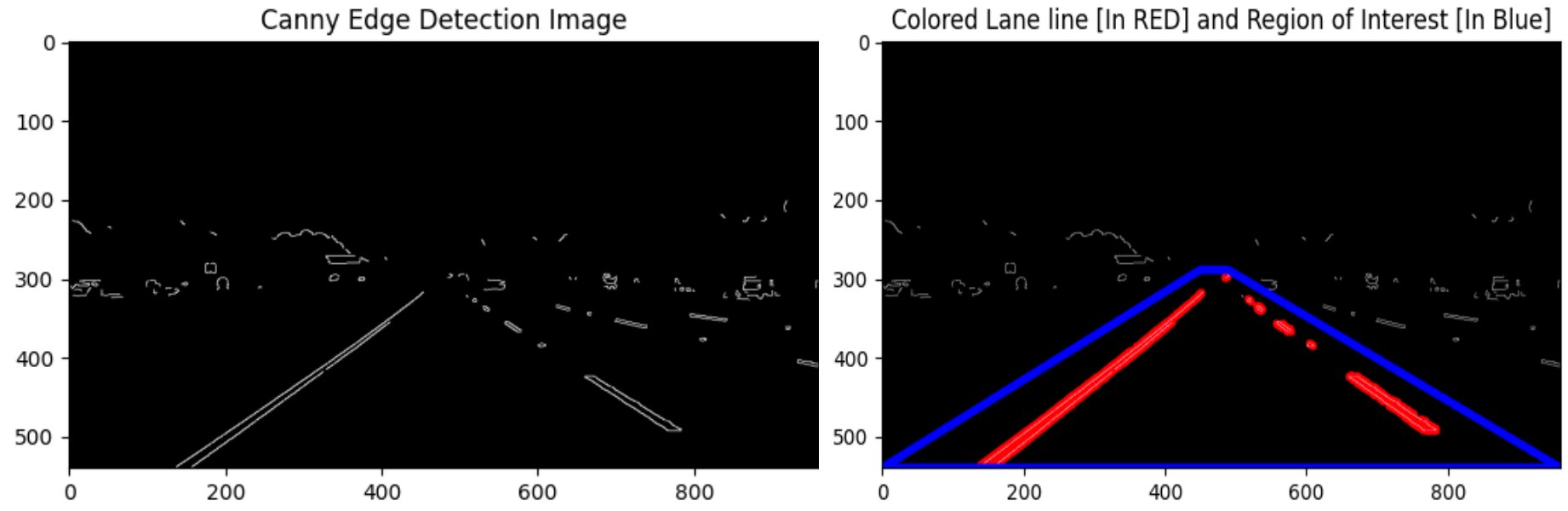

Fig. 22 Output Prediction of Lanes After Canny Edge Detection

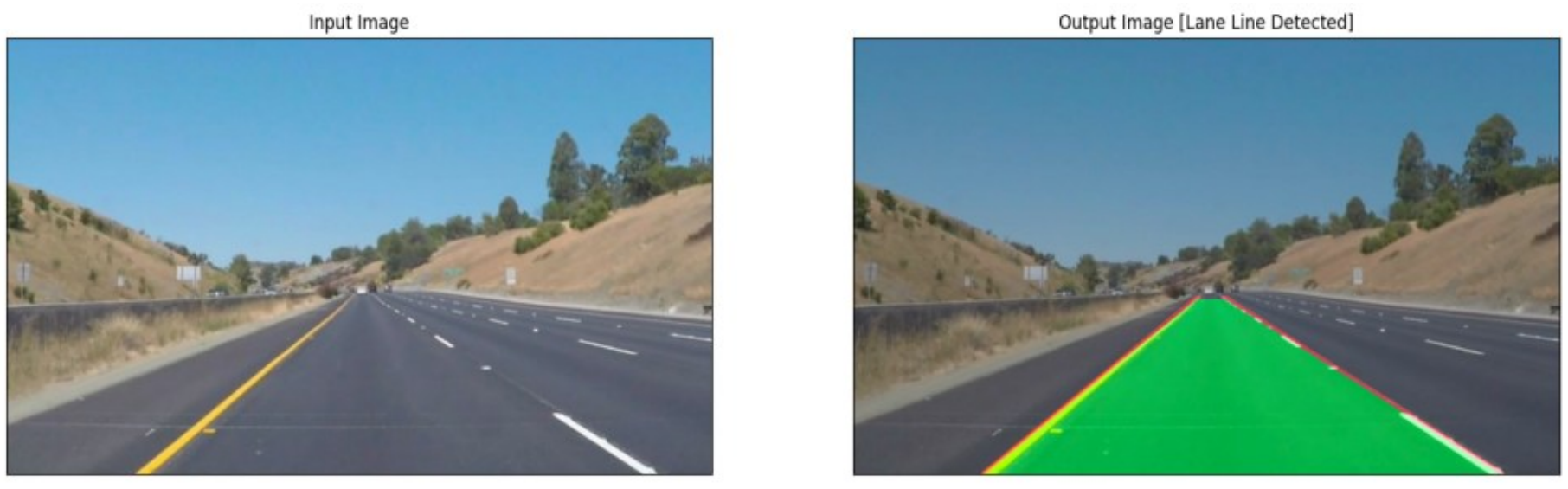

Fig. 23 Output Prediction of Lanes After Passing Through a Pipeline of Transforms

## 4.3 Vehicle Detection:

- InceptionV3:

  The datasets were resized to 75 x 75 to be given as input to the model. The weights were downloaded from ImageNet and the top layer of InceptionV3 was removed. The output of the modified inceptionv3 model was flattened in order to be passed into a dense layer of 256 neurons and ReLU activation function followed by a dropout layer of 10% and a final dense layer with 1 neuron and sigmoid activation function. The model was trained for 20 epochs with a batch size of 32, RMSprop optimizer, binary cross entropy loss function, and learning rate of 0.0001. The modified InceptionV3 architecture diagram is shown in Fig. 24.

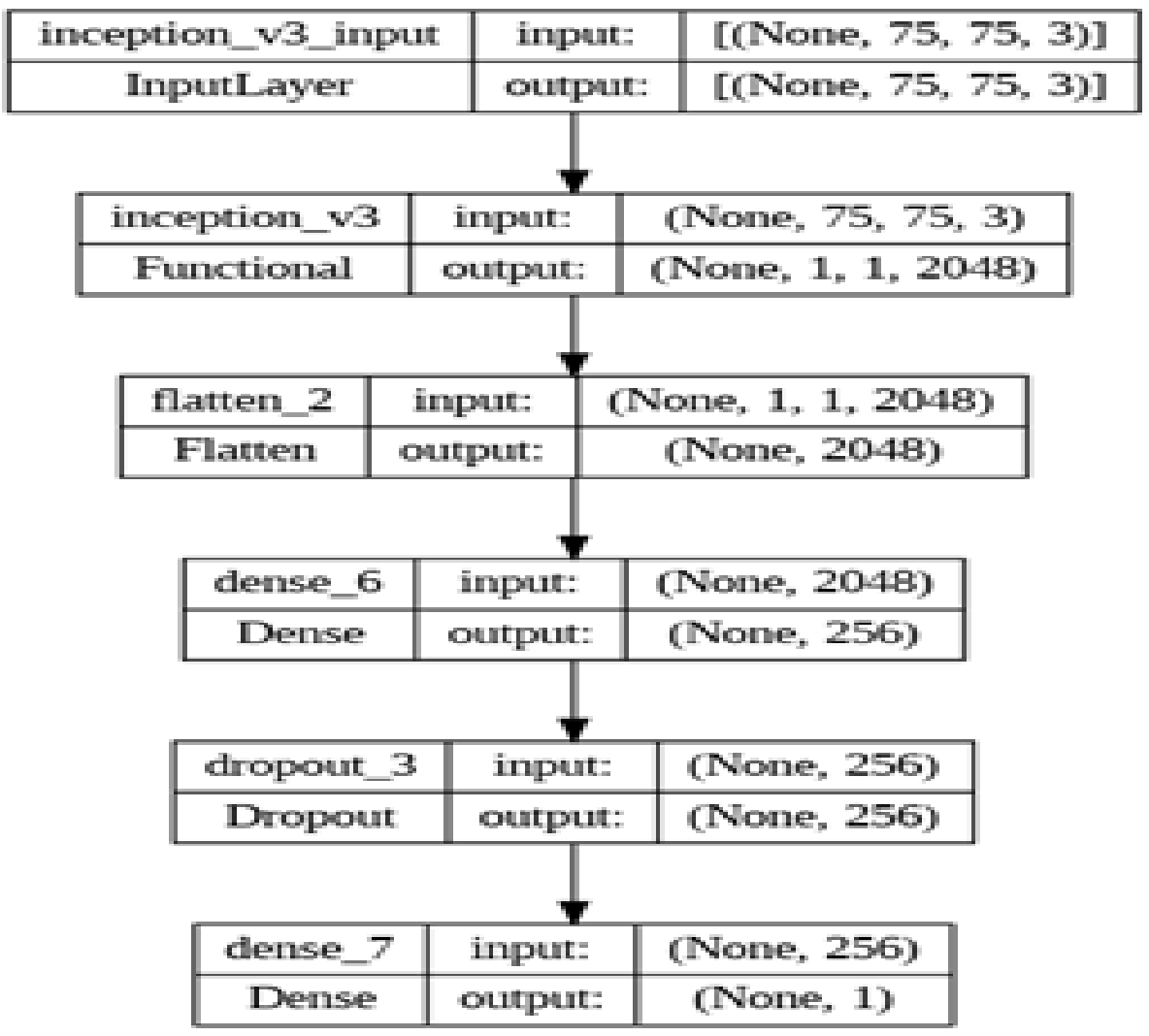

Fig. 24 Modified InceptionV3 Architecture Diagram

- Xception:

The datasets were resized to 75 x 75 to be given as input to the model. The weights were downloaded from ImageNet and the top layer of Xception was removed. The output of the modified Xception model was flattened to be passed into a 2d global average pooling layer followed by a dense layer of 128 neurons followed by a dropout layer of 10% and a final dense layer with 1 neuron and sigmoid activation function. The model was trained for 25 epochs with a batch size of 32, Adam optimizer, binary_crossentropy loss function. The modified Xception architecture is shown in Fig. 25.

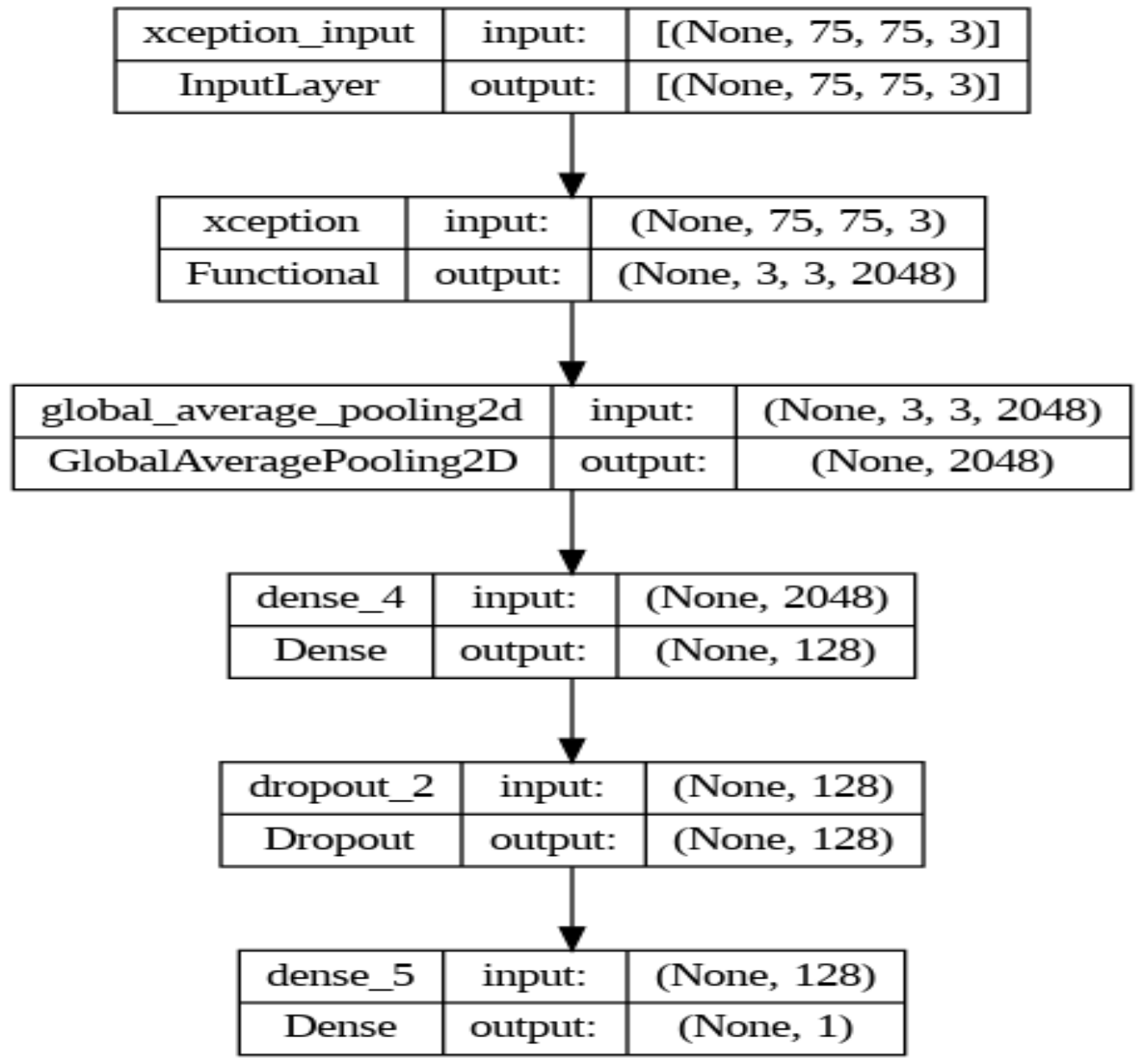

Fig. 25 Modified Xception Architecture Diagram

- MobileNet:

The datasets were resized to 75 x 75 to be given as input to the model. The weights were downloaded from ImageNet and the top layer of Mobile Net was removed. The output of the modified model was passed into a layer of Batch normalization and flattened in ordered to be passed into a 2d global average pooling layer followed by a dropout layer of 50% and a final dense layer with 1 neuron and sigmoid activation function. The model was trained for 25 epochs with a batch size of 32, Adam optimizer, binary_crossentropy loss function. The modified MobileNet architecture is shown in Fig. 26.

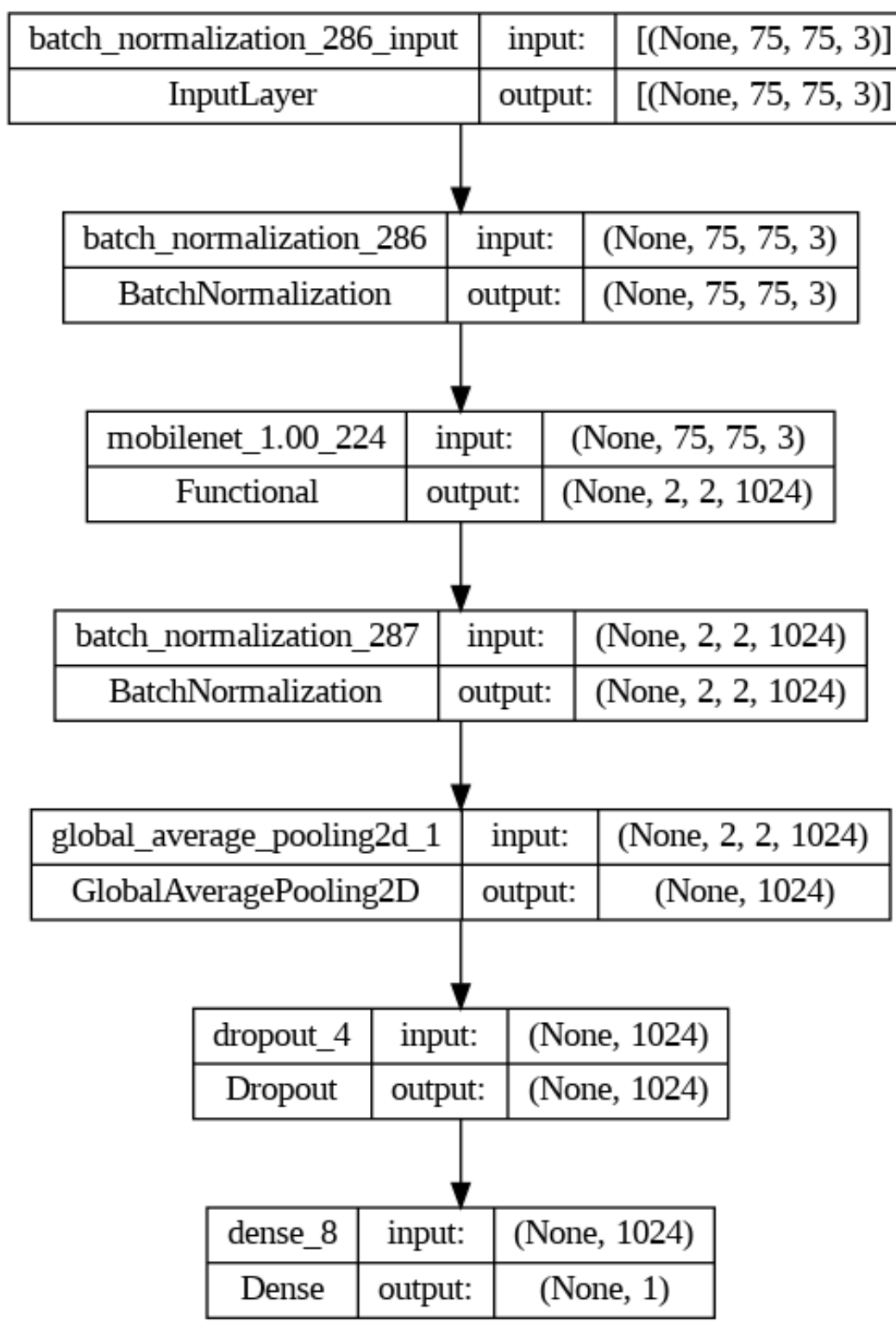

Fig. 26 Modified MobileNet Architecture Diagram

- YOLOv5:

YOLOv5 was a pre-trained model which was trained already with 80 classes of images and the model weights were stored in the format 'yolox.pt'. The yolo model can detect and classify the following classes of images: car, truck, person, etc., The model was very stable and could detect vehicles from images, videos, live webcam, etc. The refereed architecture of YOLOv5 is shown in Fig. 27. Fig. 28 shows a sample prediction of vehicle using YOLOv5.

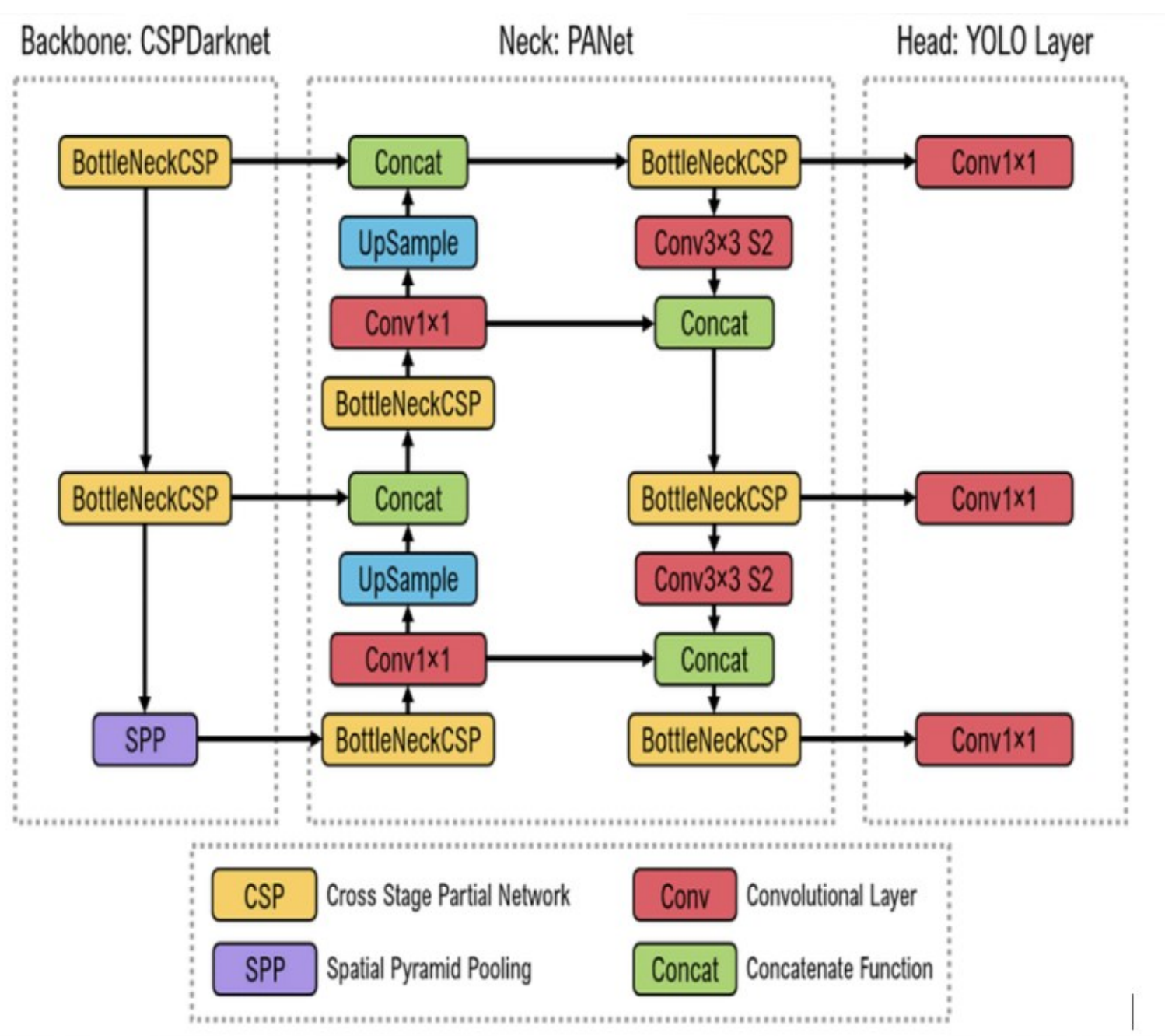

Fig. 27 YOLOv5 Architecture Diagram [36]

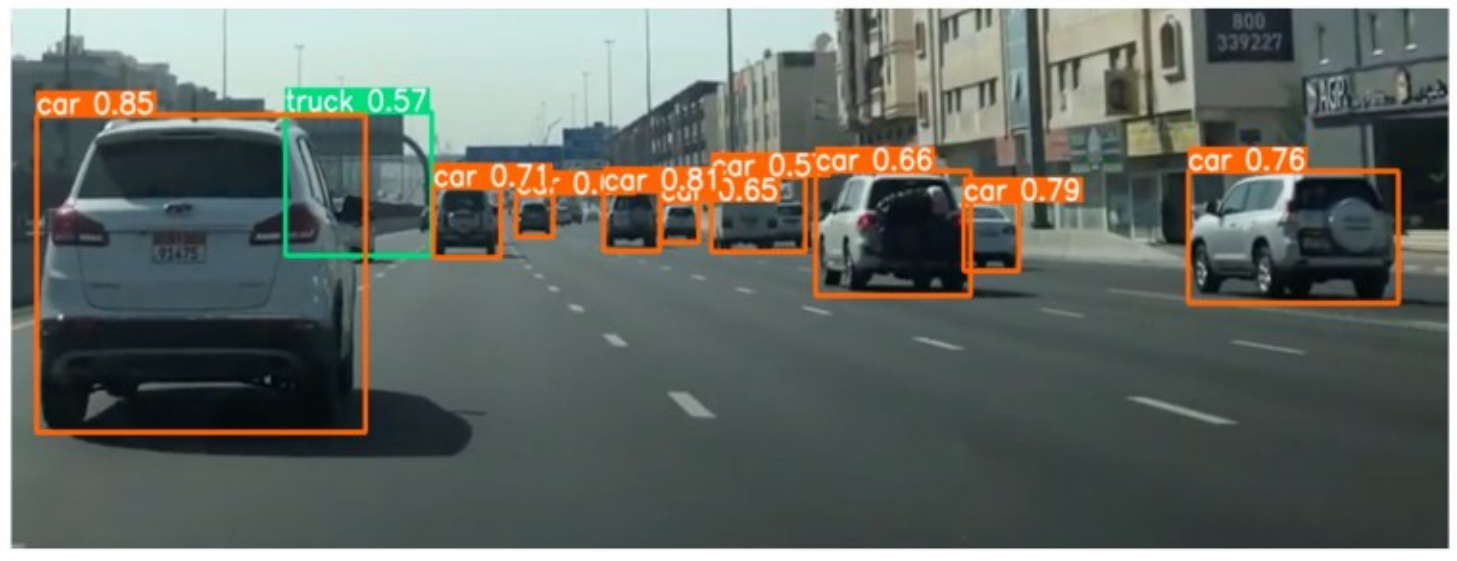

Fig. 28 Output Prediction of Vehicle by YOLOv5 Model

## 4.4 Behavior Cloning:

- ResNet50:

  The dataset was split up into train and test in the ratio 80:20 respectively. The model was trained with the help of Transfer learning using Reset50. Global average pooling was applied to the model and a dropout of 0.5 was applied. The output after dropout was flattened and passed into a Dense layer of 1 neuron with a SoftMax activation function. The dense layer is the final layer of the model and makes the prediction. The model was trained for 30 epochs with a batch size of 32, MSE loss function, and Adam optimizer with a learning rate of 0.0001. The metrics used for evaluation were accuracy and loss. The modified ResNet50 model is shown in Fig. 29.

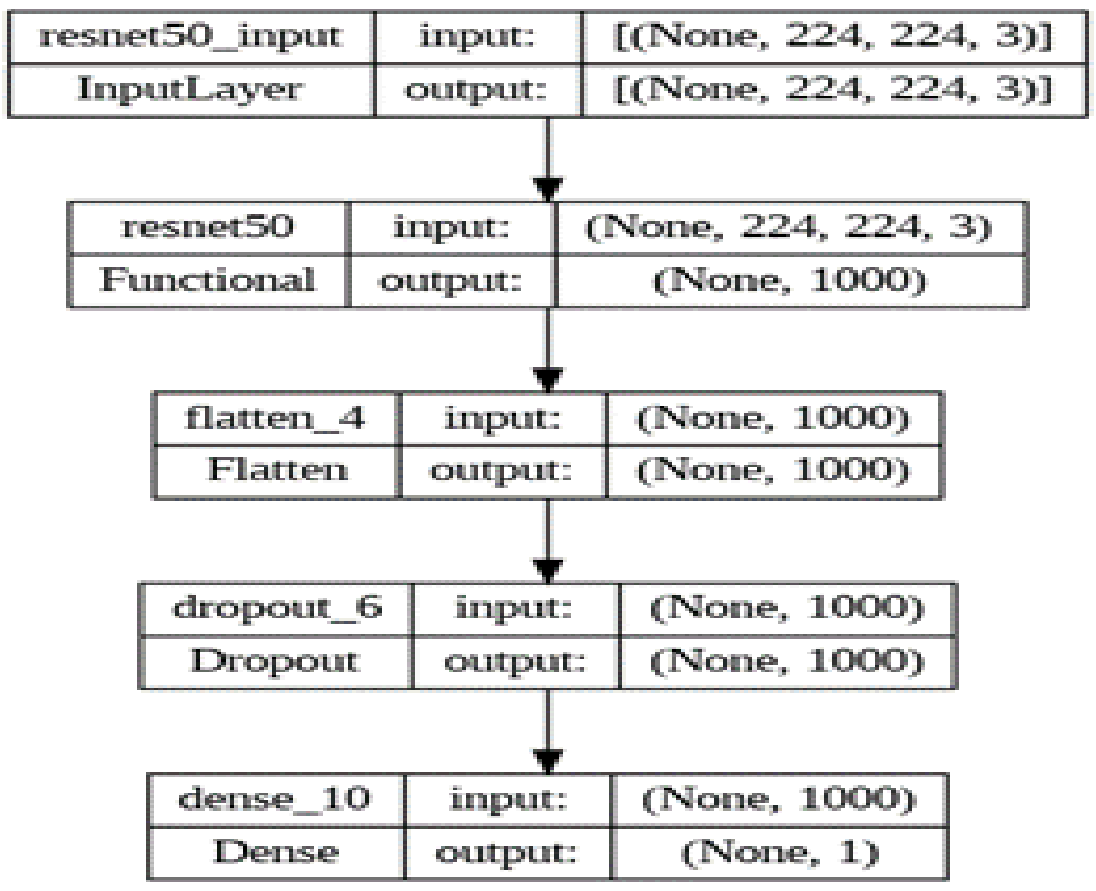

Fig. 29 Modified ResNet50 Architecture

- Proposed Model:

The images in the datasets were reshaped to the size 66 x 200. The dataset was split up into train and test in the ratio 80:20 respectively. The model was trained using a custom-built CNN. The model consisted of 5 consecutive convolutional layers with 24, 36, 48, 64, and 64 filters respectively, and a kernel size of 5 x 5 for the first 3 layers and 3 x 3 for the following 2 layers. The convolutional layers had an activation function 'ELU'. The convolutional layer was followed by a dropout layer of 50% followed by a flattening layer. The model was then connected to a dense layer of 100 neurons followed by 50,10,1 neuron respectively and the 'ELU' activation function. The model was trained for 50 epochs with a batch size of 32, Adam optimizer, and mse loss function. The metrics used for evaluation were accuracy and loss. The proposed model architecture diagram is shown in Fig. 30. Fig. 31 shows the output of behavioral cloning in the Udacity self-driving car simulator using the proposed model.

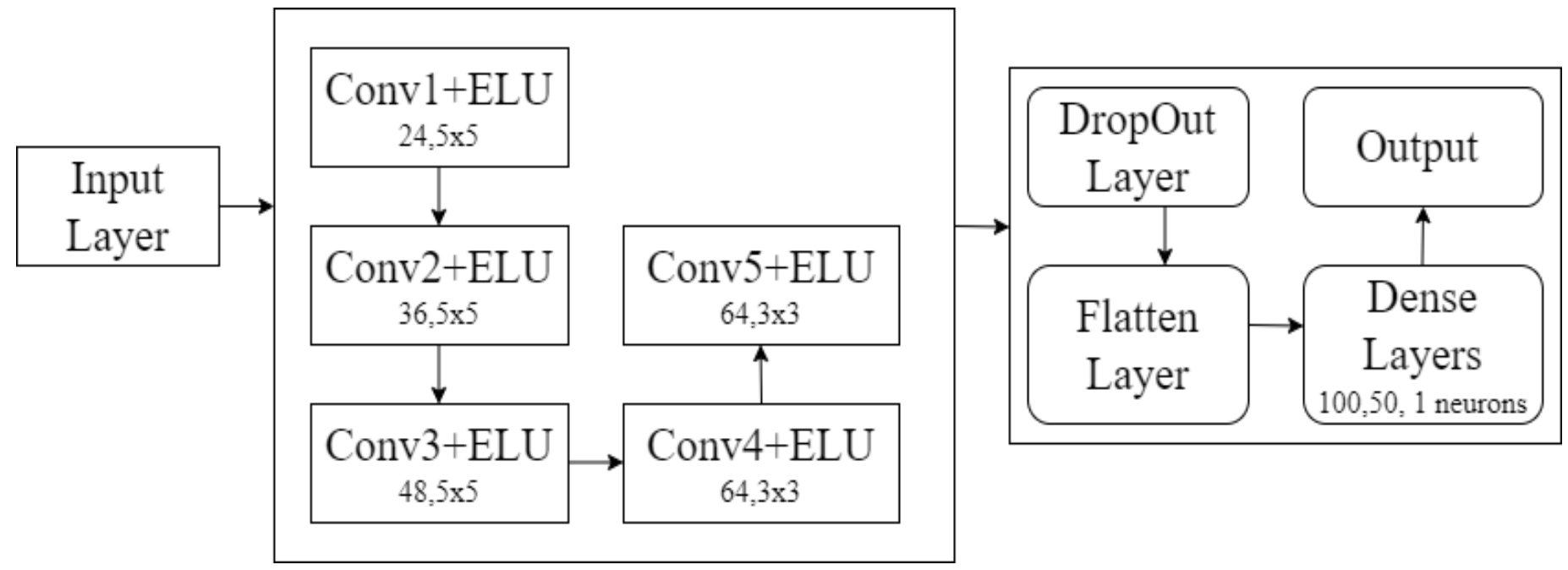

Fig. 30 Proposed Model Architecture Diagram

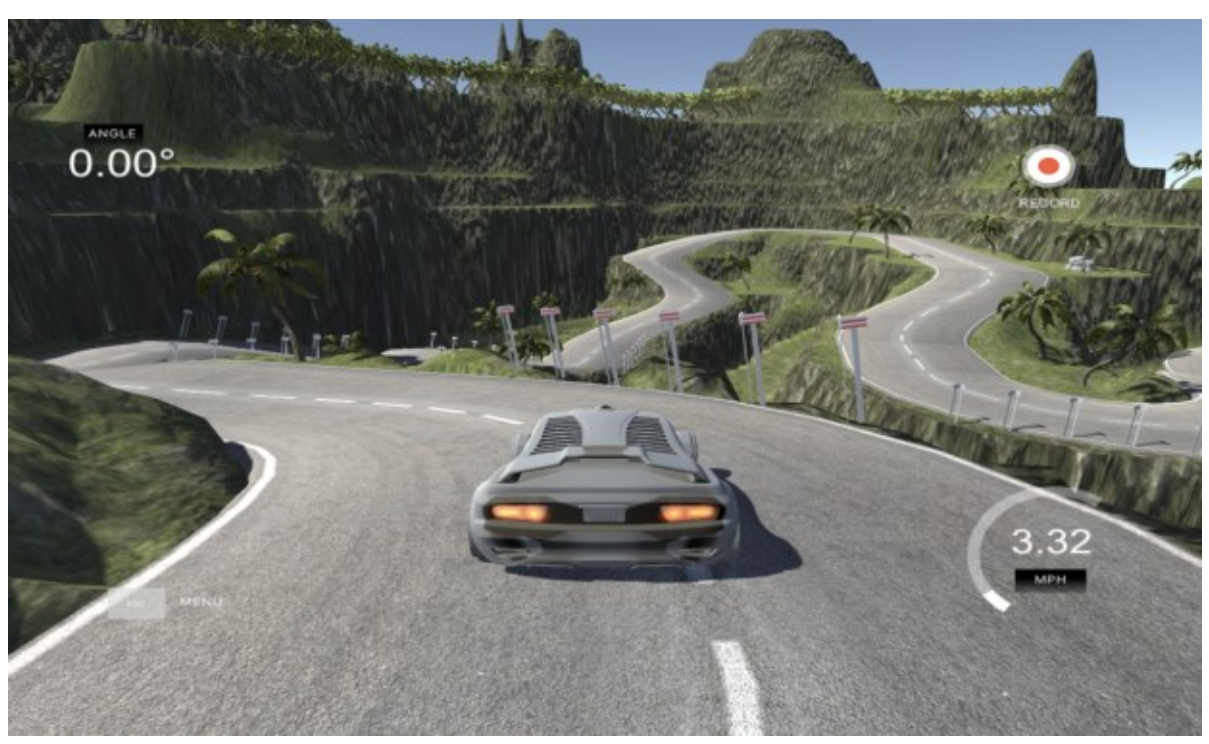

Fig. 31 Output of Behavioral Cloning in Udacity Self-driving Car Simulator

## 5. Results and Discussion

Comparative analysis of model performance between ResNet50 and a Custom CNN model was conducted for Traffic Signal Detection & Classification. The Resent50 model produced a training accuracy of 0.9942, training loss of 0.0216, Validation loss of 0.0157, and validation accuracy of 0.9956. The model had a testing accuracy of 0.9955 and a testing loss of 0.015. In contrast, the Custom CNN model produced a training accuracy of 0.9913, training loss of 0.0449, Validation loss of 0.0386, and validation accuracy of 0.9954.

The reason why ResNet50 produces much better results compared to Custom CNN is that ResNet50 is a much deeper neural network with many layers and each layer extracts more and more features. Whereas Custom CNN is a less deep network compared to ResNet50 and the number of layers for feature extractions is comparatively lesser. The number of training parameters in Custom CNN is lesser compared to Resnet50. A Custom CNN model was proposed in order to reduce the model's computational complexity, by reducing the number of layers for feature extraction. The training parameters were also reduced. The proposed Custom CNN was able to obtain an accuracy almost equal to that of the fine-tuned ResNet50 model. The proposed model was able to provide an accuracy slightly greater than in the work [37] that proposed an attention-based CNN model for traffic sign detection and almost equal but slightly low compared to the research work [38] that used a regularized CNN for detecting traffic signs. The accuracy of the ResNet model in the current research work was almost equal to the work in [38] and greater than the work in [37].

Comparative analysis of model performance between FCNN with VGG16 backbone and OpenCV Transforms was conducted for lane detection and segmentation. The FCNN Model with VGG16 backbone was trained with 3 sets of loss functions SGD, RMSprop, and Adam for comparative analysis. The FCNN model produced a training accuracy of 0.9452, training loss of 0.0616, Validation loss of 0.0357, and validation accuracy of 0.9562. The model produced the best results when trained with Adam optimizer compared to other SGD and RMSprop. SGD is the basic loss function which uses the first moment whereas RMSprop uses the second moment. But Adam uses both the first and second moment, which is the reason for better accuracy compared to the other 2 models. The FCNN model produced a validation accuracy almost equal to but slightly less than the average success rate obtained in the work [39] that proposed a CNN based lane keeping assistant system.

The model that used OpenCV Transforms produced accurate results when region masking and color selection technique was performed when the lane consisted of white lane markings. The challenge faced using this model was it was not able to detect yellow side lane lines. It was only able to detect and extract white color due to the RGB threshold. To overcome this challenge Canny Edge Detection technique was used. After undergoing RGB to grayscale conversion and Gaussian blur the edges of the image were detected irrespective of the image's color since it was converted to grayscale. Finally, the model underwent Hugh transform where the required region was specified, and the edges were detected inside the region. The pipeline was created with the process flow of transforms and each frame of the video was converted using the pipeline and produced accurate results. The challenge faced during the process was the detection of lanes during tight turns, corners, and small roads.

Comparative analysis of model performance between InceptionV3, Xception, MobileNetV2, and YOLOv5 was conducted for Vehicle Detection. The InceptionV3 model produced a training accuracy of 0.9812, training loss of 0.0505, validation loss of 0.0578, validation accuracy of 0.9859 after training for 25 epochs, and test loss of 0.0292 and test accuracy of 0.9918. Whereas the Xception model produced a training accuracy of 0.9899, training loss of 0.0346, validation loss of 0.0589, validation accuracy of 0.9859 after training for 25 epochs, and a test loss of 0.0192 and test accuracy of 0.9918. The below figures (Fig. 32, Fig. 33, Fig. 34, and Fig. 35) show the obtained results from training and validation- accuracy and loss. Fig. 36 shows the ROC curve for Inception, Xception, and MobileNet models. The models compared in the current work for vehicle detection produce a validation accuracy almost equal to the detection accuracy obtained by a CNN based model for vehicle detection during the snowy weather in the research conducted in [40].

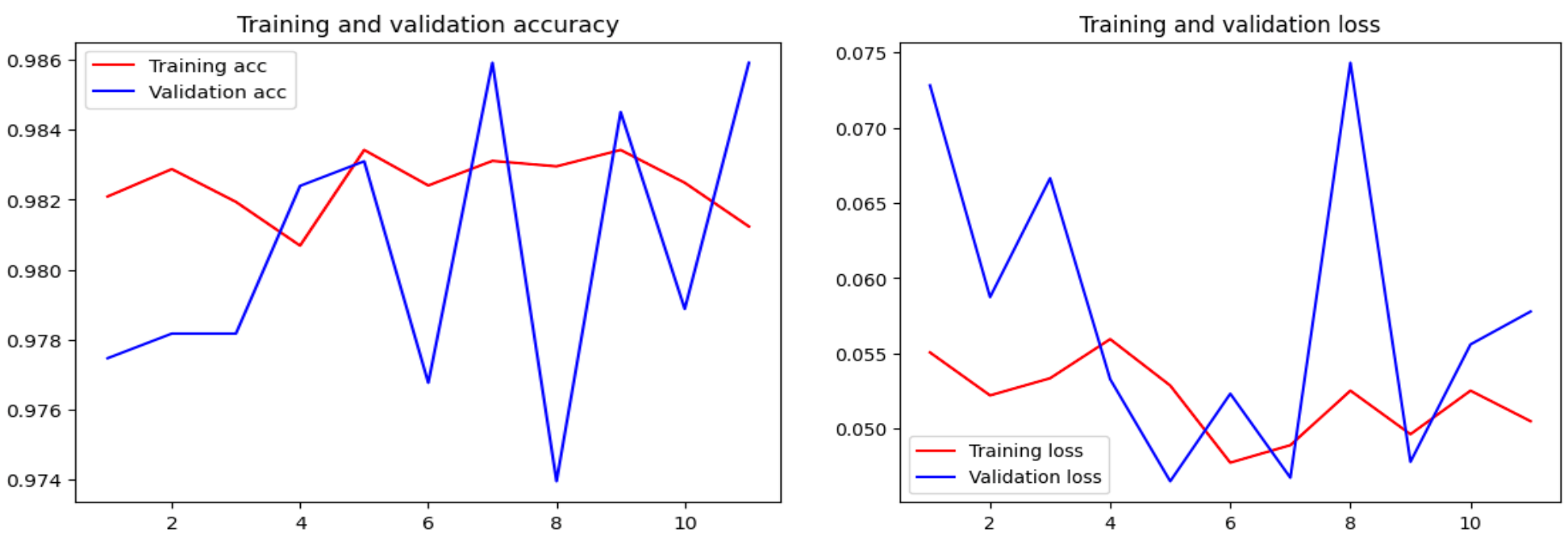

Fig. 32 Train vs Val Accuracy (InceptionV3) Fig. 33 Train vs Val Loss (InceptionV3)

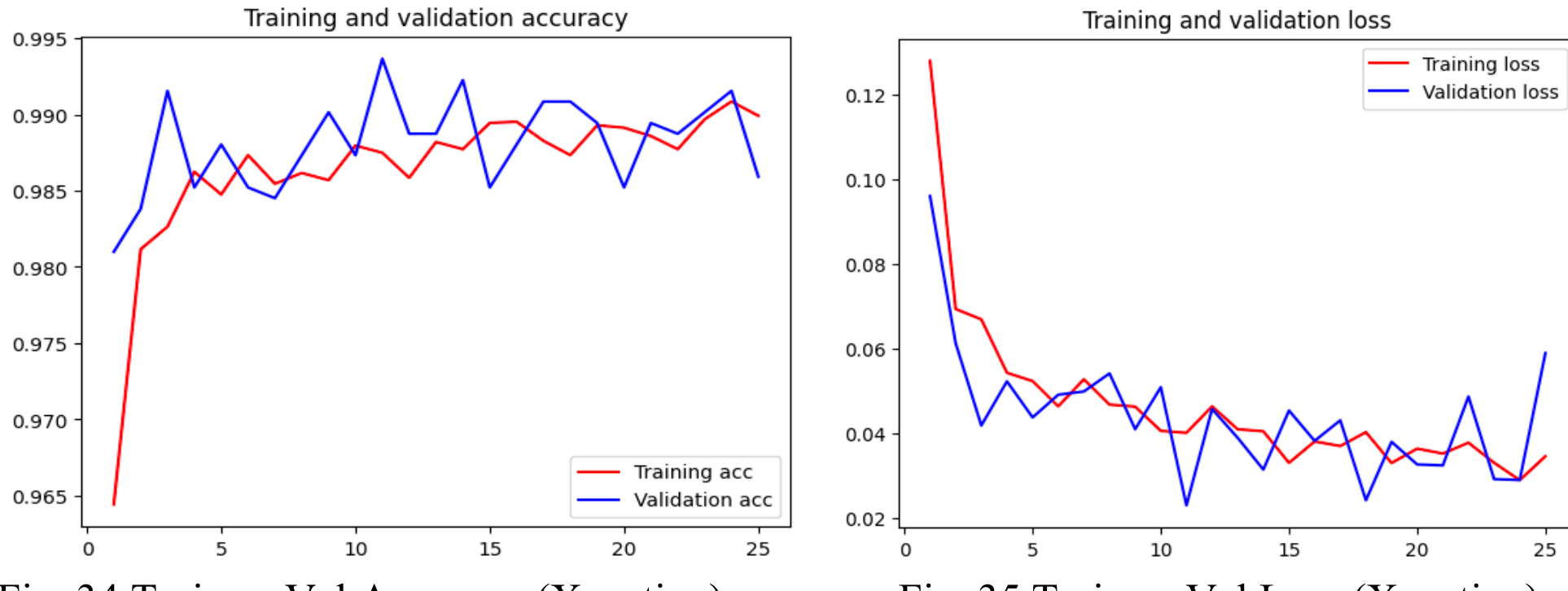

Fig. 34 Train vs Val Accuracy (Xception) Fig. 35 Train vs Val Loss (Xception)

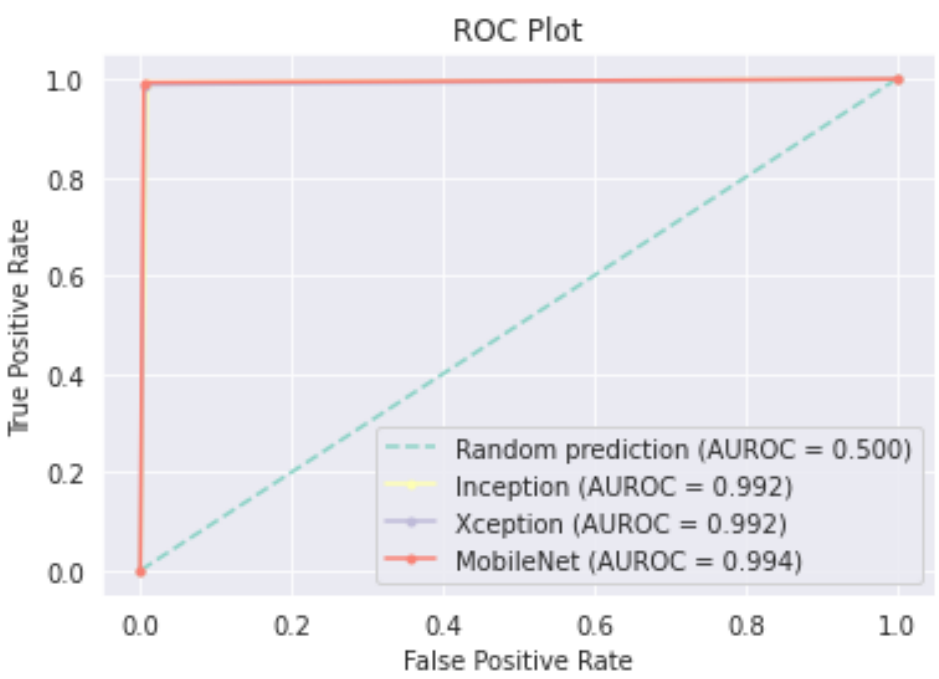

Fig. 36 ROC Curve for Inception, Xception, and MobileNet Models

The YOLOv5 model produced a very high accuracy rate and consistency. The accuracy was high while testing both images and videos. The YOLOv5 was a better model than InceptionV3 and Xception since it was able to predict a much wider range of classes, unlike InceptionV3 and Xception which could predict only vehicles or not. Yolo was a pre-trained model with more datasets and hence produced better accuracy and consistency.

Comparative analysis of model performance between ResNet 50 and Custom CNN was conducted for Behavioral Cloning. The Custom CNN model produced a training accuracy of 0.9812, a training loss of 0.1088, and a validation loss of 0.1048 after training for 30 epochs. Whereas the ResNet50 model produced a training accuracy of 0.9806, a training loss of 0.1418, and a validation loss of 0.2500 after training for 30 epochs. Fig. 37 and Fig. 38 show results for the proposed model and ResNet50. The proposed CNN and the ResNet50 model in the current study were able to provide a testing accuracy of above 98% which is higher than in the research work by [41] that used custom CNN and ResNet based models in their experimentations.

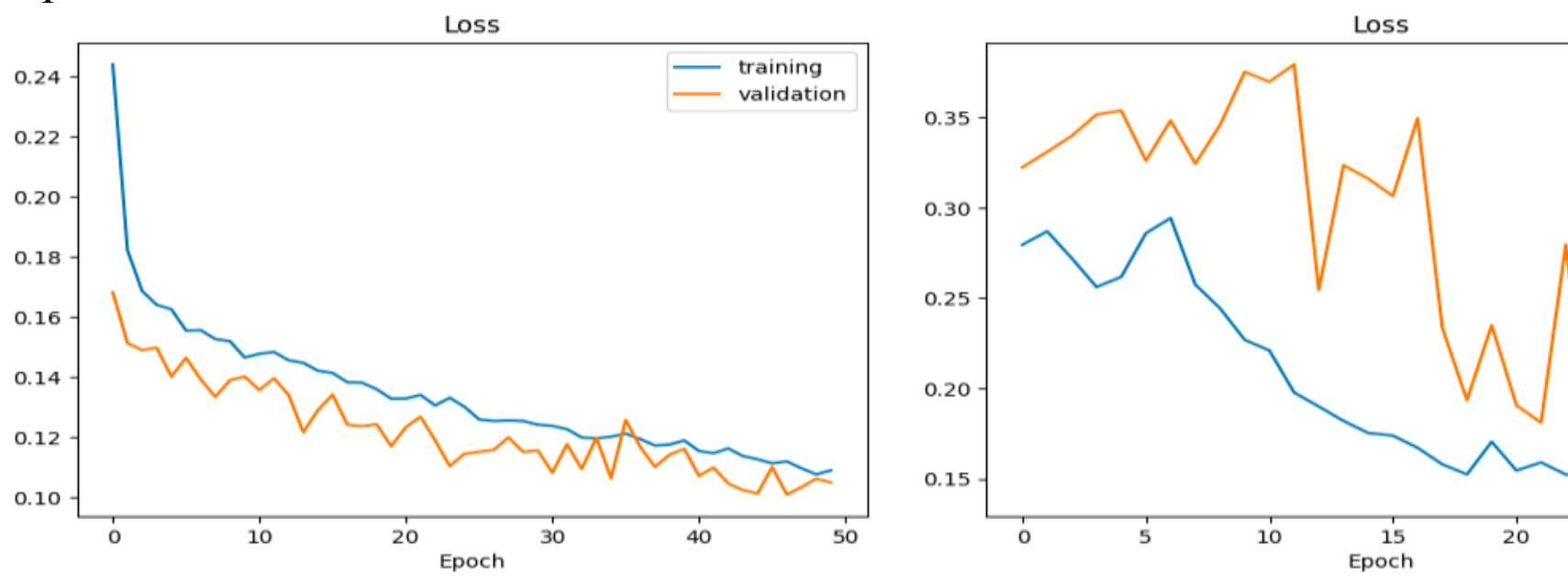

Fig. 37 Train vs Val Loss (Proposed Model) Fig. 38 Train vs Val Loss (ResNet50)

The reason why Custom CNN produces much better results compared to ResNet50 is that ResNet50 is a much deeper neural network with many layers and each layer extracts more and more features and it tends to overfit. Whereas Custom CNN is a less deep network compared to ResNet50 and the number of layers for feature extractions is comparatively lesser and it fits perfectly. The number of training parameters in Custom CNN is lesser compared to ResNet50. Furthermore, Custom CNN was trained for 50 epochs whereas the ResNet50 model was trained for 30 epochs only.

ResNet50 was very inconsistent while training which can be observed from the validation loss in the graph. The loss is more in the ResNet50 model because of something called skip layers which skips certain convolutional layers, and this has led to more inconsistent losses while training. It can also be observed that the training and validation losses don't converge together. Therefore a Custom CNN was proposed in the current study to reduce the number of training parameters, reduce inconsistency, and prevent overfitting.

For the comparative analysis and performance evaluation of the models, the following evaluation metrics were used:

$$Accuracy = \frac{TP + TN}{TP + FP + TN + FN} \tag{21}$$

$$Precision = \frac{TP}{TP + FP} \tag{22}$$

$$Recall = \frac{TP}{TP + TN} \tag{23}$$

$$F1\,Score = 2 * \frac{Precision * Recall}{Precision + Recall} \tag{24}$$

TP represents the positive samples which are correctly classified, and TN represents the negative samples which are correctly classified. FP and FN represent samples that are wrongly classified.

| | Model | Test accuracy (%) | Recall (%) | Precision (%) | F1 (%) | AUC |
|---|---|---|---|---|---|---|
| 0 | Inception | 0.992117 | 0.993135 | 0.990868 | 0.992000 | 0.992133 |
| 1 | Xception | 0.991554 | 0.988558 | 0.994246 | 0.991394 | 0.991508 |
| 2 | MobileNet | 0.994369 | 0.993135 | 0.995413 | 0.994273 | 0.994350 |

Fig. 39 Evaluation Metric Scores of InceptionV3, Xception, and MobileNet Models

Table I: Comparative Analysis of parameters of different Models

| | Model | Epochs | Batch Size | Learning rate | Image Size |
|---|---|---|---|---|---|
| **Traffic Sign Detection** | ResNet50 | 25 | 64 | 0.01 | 32x32 |
| | Custom CNN | 20 | 64 | 0.001 | 32x32 |
| **Lane Detection** | FCNN with VGG16 Backbone | 20 | 32 | 0.01 | 128x128 |
| | OpenCV Transforms | - | - | - | - |
| **Vehicle Detection** | InceptionV3 | 25 | 32 | 0.0001 | 75x75 |
| | Xception | 25 | 32 | 0.001 | 75x75 |
| | YOLOv5 | - | - | - | 640x640 |
| **Behavioral Cloning** | Custom CNN | 50 | 64 | 0.001 | 66x200 |
| | ResNet50 | 30 | 32 | 0.001 | 32x32 |

Table II: Analysis of different Models with respect to Accuracy and Loss

| | Model | Optimizer | Loss Function | Accuracy | Loss |
|---|---|---|---|---|---|
| **Traffic Sign Detection** | ResNet50 | SGD | Cross Entropy | 0.9955 | 0.015 |
| | Custom CNN | Adam | Categorical Cross Entropy | 0.9903 | 0.0439 |
| **Lane Detection** | FCNN with VGG16 Backbone | SGD | Cross Entropy | 0.4143 (Mean_io_u) | 0.1417 |
| | OpenCV Transforms | - | - | - | - |
| **Vehicle Detection** | InceptionV3 | RMSProp | Binary Cross Entropy | 0.9812 | 0.0505 |
| | Xception | Adam | Binary Cross Entropy | 0.9899 | 0.0346 |
| | YOLOv5 | Adam | Smooth L1 Loss + Binary Cross Entropy | - | - |
| **Behavioral Cloning** | Custom CNN | Adam | MSE | 0.9812 | 0.1088 |
| | ResNet50 | Adam | MSE | 0.9806 | 0.1418 |

To carry out the experiments Python 3.9.13 environment was used with TensorFlow and Keras background and it was implemented MacBook Pro M1 Silicon GPU and Google Nvidia Tesla T4 GPU. The graphs and visualizations were illustrated with the help of Matplotlib.

| Model | Time | Training accuracy (%) | Validation Accuracy (%) |
|---|---|---|---|
| Inception | 0 days 00:26:13.328518 | 0.979391 | 0.981827 |
| Xception | 0 days 00:13:25.486802 | 0.985879 | 0.988289 |
| MobileNet | 0 days 00:10:03.406209 | 0.973408 | 0.989589 |

Fig. 40 Time Analysis of InceptionV3, Xception, and MobileNet Models

Some of the areas of the current research that have to be improved to avoid failure cases are the models developed or proposed for various tasks like traffic sign detection, vehicle detection, etc, which should be incorporated with domain adaptation techniques in such a way that the performance of the proposed models does not deteriorate when evaluated under different geographical and environmental conditions.

The inference time of the proposed models can be optimized by certain techniques like model quantization or pruning for better performance in real-time applications. The system when tested to detect lanes in corners or turns failed to perform well as expected, this can be worked on by further fine-tuning the models and experimentations. The performance of the proposed system should be further improved to avoid failure cases while addressing certain challenges like unusual vehicle types or damaged traffic signs, this was not focused on during the experimentations. The performance of the developed models is slightly reduced with densely packed scenes with many vehicles or signs that slightly confuse the model. The proposed models need to be evaluated on more complex real-time applications during experimentations to avoid failure cases.

## 6. Conclusion

Through this study, we can conclude that our proposed approach effectively solves various challenges related to self-driving cars like traffic sign classification, lane prediction, vehicle detection, and behavioral cloning. The experiments performed conclude that this novel methodology can outperform several state-of-the-art methods previously used for a similar task. A comparative analysis has been performed to determine and choose the best model for a particular task.

This study still maintains the scope for further improvement and betterment by testing and training with some real-life challenging datasets and preprocessing them. The trained models can be tested in real-time with a webcam or live video feed. Instead of training with pre-trained models, convolutional neural networks can be built from scratch. Experiments can be conducted to fine-tune the hyperparameters and models can be trained for more epochs. The trained models can alternatively be tested out in Carla virtual car environment simulator. Research on challenging problems like lane detection in corners and turns can be worked on in the future. Additionally, a distance measurement algorithm can be developed to predict the distance between nearby vehicles while traveling.

## Declarations

- Funding: Not applicable.
- Conflicts of interest=Competing interests: Not applicable.
- Availability of data and material: The dataset generated during and=or analyse during the current study are available from the corresponding author.
- Code availability: Not applicable